\documentclass{article}

\usepackage{microtype}
\usepackage{graphicx}
\usepackage{subfigure}
\usepackage{booktabs} 

\usepackage{hyperref}

\usepackage[archival]{icml2023}

\usepackage{amsmath}
\usepackage{amssymb}
\usepackage{mathtools}
\usepackage{amsthm}

\usepackage[capitalize,noabbrev]{cleveref}

\usepackage[normalem]{ulem} 
\usepackage{enumitem}
\theoremstyle{plain}
\newtheorem{theorem}{Theorem}[section]
\newtheorem{proposition}[theorem]{Proposition}
\newtheorem{lemma}[theorem]{Lemma}

\theoremstyle{definition}

\newtheorem{assumptions}{Assumptions}[section]
\theoremstyle{remark}

\usepackage{stmaryrd}
\usepackage{diagbox}

\providecommand{\expt}[1]{\langle #1 \rangle}
\providecommand{\exptbig}[1]{\Big\langle #1 \Big\rangle}
\providecommand{\abs}[1]{\lvert #1 \rvert}

\providecommand{\norm}[1]{\lVert #1 \rVert}

\newcommand{\bbR}{\mathbb{R}}
\newcommand{\cB}{\mathcal{B}}
\newcommand{\cD}{\mathcal{D}}
\newcommand{\cK}{\mathcal{K}}
\newcommand{\cL}{\mathcal{L}}
\newcommand{\cN}{\mathcal{N}}
\newcommand{\cM}{\mathcal{M}}
\newcommand{\cO}{\mathcal{O}}
\newcommand{\cT}{\mathcal{T}}
\newcommand{\cU}{\mathcal{U}}
\newcommand{\cV}{\mathcal{V}}

\renewcommand{\v}[1]{\mathbf{#1}}
\newcommand{\inpd}[2]{\langle #1\,,\, #2\rangle} 
\newcommand{\inpdmid}[2]{\big\langle #1\,,\, #2\big\rangle} 
\newcommand{\pp}[2]{\frac{\partial#1}{\partial#2}}
\newcommand{\dd}[1]{\mathrm{d}#1}

\usepackage[textsize=tiny]{todonotes}

\icmltitlerunning{Linear Regression on Manifold Structured Data: the Impact of Extrinsic Geometry on Solutions}

\begin{document}

\onecolumn
\icmltitle{Linear Regression on Manifold Structured Data:\\the Impact of Extrinsic Geometry on Solutions}



\icmlsetsymbol{equal}{*}

\begin{icmlauthorlist}
\icmlauthor{Liangchen Liu}{uta}
\icmlauthor{Juncai He}{kst}
\icmlauthor{Richard Tsai}{uta,oden}
\end{icmlauthorlist}

\icmlaffiliation{uta}{Department of Mathematics, University of Texas at Austin, Austin, TX, USA}
\icmlaffiliation{oden}{Oden Institute for Computational Engineering and Sciences, University of Texas at Austin, Austin, TX, USA}
\icmlaffiliation{kst}{ Computer, Electrical and Mathematical Science and Engineering Division,  King Abdullah University of Science and Technology, Thuwal, Saudi Arabia }

\icmlcorrespondingauthor{Liangchen Liu}{lcliu@utexas.edu}

\icmlkeywords{Applied Differential Geometry, Local Linear Regression, Data Manifolds, Machine Learning}

\vskip 0.3in
%



\printAffiliationsAndNotice{}  

\begin{abstract}
  In this paper, we study linear regression applied to data structured on a manifold. We assume that the data manifold is smooth and is embedded in a Euclidean space, and our objective is to reveal the impact
  of the data manifold's extrinsic geometry on the regression. Specifically, we analyze the impact of the manifold's curvatures (or higher order nonlinearity in the parameterization when the curvatures are locally zero) on the uniqueness of the regression solution.
Our findings suggest that the corresponding linear regression does not have a unique solution when the embedded submanifold is flat in some dimensions. Otherwise, the manifold's curvature (or higher order nonlinearity in the embedding) may contribute significantly, particularly in the solution associated with the normal directions of the manifold. Our findings thus reveal the role of data manifold geometry in ensuring the stability of regression models for out-of-distribution inferences.
\end{abstract}


\section{Introduction}
The manifold hypothesis posits that real-world data points typically cluster on a lower-dimensional manifold, denoted as $\mathcal M$, within a high-dimensional encoding space like $\mathbb{R}^d$.
This concept has been investigated in numerous studies \cite{hein2006manifold, narayanan2010sample, niyogi2008finding,tenenbaum2000global}.  Fefferman et al. \yrcite{fefferman2016testing} established a theoretical and algorithmic framework to statistically validate and test the manifold hypothesis. Other rigorous experiments~\cite{brand2002charting, ruderman1994statistics,roweis2000nonlinear,scholkopf1998nonlinear} have also provided empirical evidence supporting the existence of low-dimensional manifold, particularly in the context of image data sets.

The related current mainstream research fields can be classified into the following two categories: (i) Dimensionality reduction methods~\cite{abdi2010principal, roweis2000nonlinear,tenenbaum2000global};
(ii) Approximation theory for deep learning models for functions supported on low-dimensional manifolds~\cite{chen2019efficient,cloninger2021deep,shaham2018provable}.
One of the objectives of (i) is to reduce the ``curse of dimensionality" imposed by the embedding space directly and explicitly by (approximately) encoding the data into a lower dimensional space.
In the case of (ii), results show that neural networks can approximate functions that are defined on an embedded smooth manifold with costs depending on the intrinsic dimensions of the manifold itself not the dimensionality of the embedding space.
However, these studies use only some global geometric quantities, such as the reach~\cite{federer1959curvature} of the data manifold to establish the approximation estimates.
Moreover, relatively little is known about the stability of neural networks when evaluating inputs that deviate from the manifold, i.e., inputs that lie outside the training data distribution. Further research is needed to uncover the stability characteristics of neural networks in these situations and extend our knowledge of their performance and limitations.

Instead of studying the dimensionality reduction or approximation techniques when data are concentrated on a low-dimensional manifold, we propose to explicitly explore the influence of extrinsic geometric information of the data manifold on the learning process and outcome.
In particular, we study linear regression models under the assumption that the data is distributed on a smooth manifold embedded in a higher dimensional Euclidean space. To unveil the influence of the manifold's geometry on the regression, we focus our analysis on a localized region {around} the manifold. We thus refer to linear regression in such a setup as the local linear regression problem.
Specifically, we focus on the issues of 
uniqueness and stability of solutions in the well-posedness of the linear regression, stemming from the local geometry of the data manifold.
We further discuss the effect of adding noise as regularization for the local linear regression under the low-dimensional manifold setup.

More precisely, we assume a smooth low-dimensional data manifold $\mathcal M$ embedded in $\bbR^d$, and we have the data/target function $g:\bbR^d\to\bbR$. Local linear regression means that we construct an affine function
\begin{equation}\label{eq:affine_f}
    f(\v x,\v w, b) = \v w\cdot\v x + b,~~~\v x,\v w\in\mathbb{R}^d,b\in\mathbb{R},
\end{equation}
by solving the least square optimization problem
\begin{equation} \label{eqn:leastsquareloss}
    \min_{\v w\in\mathbb{R}^{ d}, b\in\mathbb{R}}\cL(\v w, b) = \int_{\Omega} \abs{f(\v x,\v w, b) - g(\v x)}^2 \rho(\v x)\dd{\v x},
\end{equation}
where $\rho(\v x)$ is the data density in a small neighborhood $\Omega\subset \mathcal{M}$.
The local linear regression model is of interest for two primary reasons.
On the one hand, the local linear regression model, which is also known as the moving least-square method~\cite{cleveland1979robust,cleveland1988locally,levin1998approximation}, is widely studied in computer graphics~\cite{schaefer2006image}, numerical analysis~\cite{gross2020meshfree,liang2013solving}, and machine learning~\cite{trask2019gmls, wang2010moving}.
On the other hand, any deep neural network (DNN) with ReLU activation function is essentially a piece-wise linear function~\cite{arora2016understanding}, which can be interpreted as a different parametrization of the local linear regression model or the linear adaptive finite element method~\cite{babuvvska1978error}.
By further studying the approximation theory of adaptive finite elements and connections between ReLU DNNs and the linear finite element methods~\cite{he2020relu}, one can show that the linear region of the learned ReLU DNN is relatively small if the Hessian of the target function has a lower bound on that region.
Consequently, the optimal learned ReLU DNN should comprise multiple local linear regressions within small linear regions if the Hessian of the target function is not small.
In addition to these, a general regression model $f(\v x; \theta) = \v w \cdot \psi(\v x;\tilde \theta) + b$ with feature map $\psi(\v x;\tilde \theta)$ can be understood as a linear regression on the latent manifold $\psi\left(\mathcal M\right) := \{ \psi(\v x;\tilde \theta) \,:\, \v x \in \mathcal M\}$. Therefore, studying how the geometric information of the manifold affects the linear regression system can benefit us in understanding the general regression model or regularizing the feature map~\cite{zhu2018ldmnet}.

Moreover, it is important to consider that datasets in practice often contain noise. Thus, it is imperative to delve into the impact of noises on the linear regression solution.
As studied in~\cite{he2023side}, noise in the codimension of the data manifold
can provide a regularization that improves the stability of linear or ReLU DNN regressions. In this work, we further showcase how the presence of noise can potentially prevent the degeneracy and provide regularization effects for the linear regression model when the embedded data manifold is not flat. 

{To summarize}, in this study, we focus on local linear regression models with data on low-dimensional submanifolds embedded in a Euclidean space and mainly investigate the following questions:
\begin{enumerate}
    \item How the uniqueness of solutions for the local linear regression problem is dependent on the local geometric information, such as curvatures, of the data manifold;
    \item How the local geometric information affects the regression outcomes;
    \item How noises interact with the local geometric information of the data manifold and affect learning.
\end{enumerate}

\subsection*{Other related work}
\paragraph{Linear dimension reduction, manifold learning, and the intrinsic dimensionality of data and features.}

A multitude of dimensionality reduction methods exists, both in the supervised and unsupervised settings, including
Linear Discriminant Analysis (LDA)~\cite{balakrishnama1998linear}, Principal Component Analysis (PCA)~\cite{abdi2010principal}, Multiple Dimensional Scaling (MDS)~\cite{cox2008multidimensional}, and Canonical Correlation Analysis (CCA)~\cite{hardoon2004canonical}.
The random projection framework provides a theoretical justification for data compression~\cite{bourgain2011explicit,johnson1984extensions,krahmer2011new} using random matrices and sampling methods.
Manifold learning algorithms~\cite{belkin2003laplacian,brand2002charting,chui2018deep,donoho2003hessian,roweis2000nonlinear, saul2003think, tenenbaum2000global, weinberger2004learning}, as a direct result of low-dimensional manifold hypothesis, aims at finding local low dimensional representations of the high dimensional data.
In the context of deep learning, Gong et al.~\yrcite{gong2019intrinsic} finds the intrinsic dimensionality of deep neural network representations is significantly lower than the dimensionality of the embedded space of data. Across layers of neural networks, Ansuini et al.~\yrcite{ansuini2019intrinsic} further showcases that the intrinsic dimension of features first increases and then progressively decreases in the final layers.
Pope et al.~\yrcite{pope2021intrinsic} investigates the role of low-dimensional structure in deep learning by applying dimension estimation tools to natural image datasets. They find that neural networks could learn and generalize better on low-dimensional datasets.

\paragraph{Approximation of functions supported on a manifold.}
Several studies have shown that the approximation rate for functions defined on low-dimensional manifolds is determined by the intrinsic dimensions of the manifolds rather than the dimensions of the ambient spaces.
Shaham et al.~\cite{shaham2018provable} achieves this rate by utilizing the wavelet structure to construct a sparsely-connected neural network.
Chen et al.~\yrcite{chen2019efficient} and Schmidt~\yrcite{schmidt2019deep} implement chart determination (or partition of unity) and Taylor approximation with ReLU neural network to get the results.
Liu et al.~\yrcite{liu2021besov} obtains similar approximation rates for approximating Besov functions with convolutional residual networks.   
As for stability, a more comprehensive understanding requires making specific assumptions about the form of the functions being approximated. For instance, Cloinger et al.~\yrcite{cloninger2021deep} explored the case where functions are locally constant along the normals of the manifold.

\paragraph{The impact of the data manifold in learning.}
The exploration of how the geometry of the data manifold affects the learning process and how to utilize this information to enhance learning outcomes has received limited attention in the existing literature. For deep linear and ReLU neural networks, He et al.~\yrcite{he2023side} investigates the variation of the learned functions in the direction transversal to a linear subspace where the training data is distributed. This work also delves into the side effects and regularization properties of network depth and noise within the codimension of the data manifold.
Following the low-dimensional manifold hypothesis, Zhu et al.~\yrcite{zhu2018ldmnet} proposes to apply the dimensionality of the data manifold as a regularizer in deep neural networks to achieve better performance for image classification. Furthermore, Dong et al.~\yrcite{dong2020cure} uses the curvature information as the regularizer for missing data recovery tasks.

\section{Local linear regression on embedded manifolds}
\label{sec:regression_formula_manifold}



Denote $\v x=(x_1,x_2,\cdots,x_d)$ and $\v w=(w_1,w_2,\cdots,w_d)$.
The solution to the linear regression problem \eqref{eq:affine_f}-\eqref{eqn:leastsquareloss}
should satisfy
the following $(d+1)\times (d+1)$ linear system, which is also known as the least square problem:
\begin{equation}\label{eqn:lin_systm}
\begin{bmatrix}
		\expt{x_1^2} & \expt{x_1x_2}& \dots & \expt{x_1x_d}&\expt{x_1}\\
		\expt{x_1x_2}& \expt{x_2^2}&\dots & \expt{x_2x_d} &\expt{x_2}\\
		\vdots & \quad & \ddots & \quad & \vdots\\
		\expt{x_1x_d} & \expt{x_2x_d}  & \dots & \expt{x_d^2}&\expt{x_d} \\
		\expt{x_1} & \expt{x_2}  & \dots & \expt{x_d}& 1
	\end{bmatrix}\begin{bmatrix}
			w_1 \\ w_2 \\ \vdots \\ w_d \\ b
\end{bmatrix}= \begin{bmatrix}
	\expt{gx_1} \\ \expt{gx_2} \\ \vdots \\ \expt{gx_d} \\ \expt{g}
\end{bmatrix},
\end{equation}
where the $\expt{\,\boldsymbol{\cdot}\,}$ notation denotes averaging with the density $\rho(\v x)$: 
\[\expt{\,\boldsymbol\cdot\,} = \int_\Omega\;\boldsymbol\cdot\; \rho(\v x)\dd{\v x}.\]
We say that the linear regression problem \eqref{eq:affine_f}-\eqref{eqn:leastsquareloss} is ill-posed when \eqref{eqn:lin_systm} does not have a unique solution. 

{When the scope of the linear regression is restricted to a local subset of the data manifold $\cM$, the corresponding analysis can be simplified through a suitable change of coordinates given by a unitary transformation followed by a translation. The following simple lemma shows that this change of coordinates preserves the equivariance of the linear regression problem \eqref{eq:affine_f}-\eqref{eqn:leastsquareloss}.}

\begin{lemma}
 \label{thm:invariant}
 Let  $Q\in \mathbb{R}^{d\times d}$ be an orthogonal matrix and $\mathbf{t}_0\in\mathbb{R}^d.$ If $(\mathbf{w^*}, b^*)\in \mathbb{R}^d\times\mathbb{R}$ minimizes
\[
\cL(\v w, b) := \int_\Omega \abs{(\mathbf{x}\cdot\v w+ b) - g(\v x)}^2 \rho(\v x)\dd{\v x}
\] 
then
$(Q \v w^*,\, b^*- (Q\v w^*)\cdot \v t_0)$ minimizes 
\[
 \tilde{\cL}(\v w, b) := \int_\Omega \abs{(Q\mathbf{x}+\v t_0)\cdot\v w+ b - g(\v x)}^2 \rho(\v x)\dd{\v x}
\] 
\end{lemma}

\begin{proof}
	Define the affine transformation $\phi( \v x) = Q\v x + \v t_0$  and denote $\widetilde{\v x}=\phi(\v x)$ and $\tilde\Omega$ be the image of $\Omega$ under this affine transformation. Since $Q$ is orthogonal and has full rank, $\phi$ is a change of coordinates. Let 
     $\widetilde{\v w} = (\widetilde{w}_1,\, \widetilde{w}_2,\,\dots,\,\widetilde{w}_{d-1},\,\widetilde{w}_y)$ and $\widetilde b$ be the new set of parameters for the linear regression problem in the transformed domain. Further define $\widetilde g(\widetilde{\v x})= g\big (  Q^T(\widetilde{\v x}-\v t_0)\big )$ so that $\tilde g\big (\phi(\v x)\big )=g(\v x)$, and similarly~$\tilde \rho$. 
	\begin{align*}
			&\min_{\widetilde{\v w}, \tilde b} \cL(\widetilde{\v w},\, \widetilde b)  
			 = \min_{\widetilde{\v w}, \widetilde b} \int_{\widetilde{\Omega}} \abs{\widetilde{\v w}^T \widetilde{\v x} - \tilde g(\widetilde{\v x})}^2 \tilde\rho(\widetilde{\v x})
             \dd{\tilde{\v x}}\\
			  =& \min_{\widetilde{\v w}, \widetilde b} \int_{\Omega} \abs{\widetilde{\v w}^T\v t_0+\widetilde{\v w}^TQ\v x+\widetilde b - g(\v x)}^2 \rho({\v x})|det(Q)|\dd{\v x}\\
			   =& \min_{\v w, b} \int_{\Omega}  \abs{\v w^T\v x + b - g(\v x)}^2 \rho({\v x}) \dd{\v x} = \min_{\v w, b} \cL(\v w,\, b).
	\end{align*}
	Above, we use the fact that $|det(Q)| = 1$ since $Q$ is orthogonal.
 Therefore, the equivalence between the minimization problems implies the minimizer of the transformed problem $\{\widetilde{\v w}, \tilde b\}$ can be identified with that of the original problem $\{\v w, b\}$  through the following bijection:
$
\v w^T = \widetilde{\v w}^TQ, \; b = \widetilde{\v w}^T\v t_0 + \tilde b.
$
\end{proof}

\begin{figure}[H]
    \centering
    \includegraphics[width=0.5\textwidth]{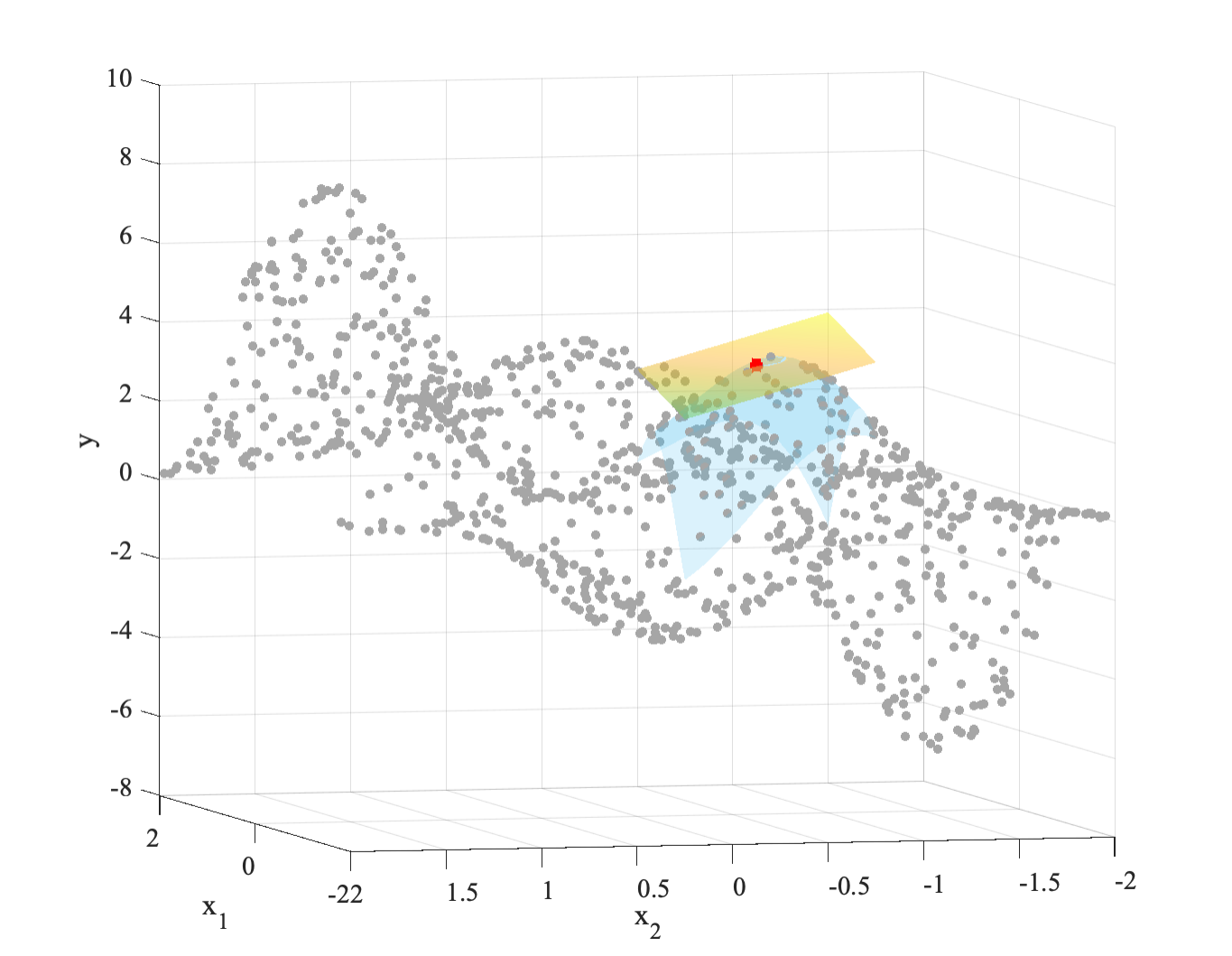}
    \caption{A visualization of a collection of manifold-structured data (the grey point clouds) sampled from a $2$-manifold embedded in $\bbR^3$; the local based point $\v x_0$ is given by the red dot; the corresponding tangent space $\cT_{\v x_0}\cM$ is the yellow plane; the blue surface indicates the part of the submanifold that would be used for local linear regression. Note only the data distribution is shown here, no information of the target function $g$ is presented.}
    \label{fig:manifold_demo}
\end{figure}

So, without loss of generality, we study the least square problem under a suitable local coordinate system, which can facilitate the following queries related to local linear regression on embedded submanifolds. 
\begin{enumerate}
    \item Examining the solvability of \eqref{eqn:lin_systm};
    \item Deriving explicit solution formulas;
    \item Describing the effect of the data manifold's geometry on local linear regression more lucidly.
\end{enumerate}
Finally, we outline the assumptions and conventions that will be employed throughout the rest of this work.
\begin{assumptions} We have the following assumptions for the data and model with a visual provided in \Cref{fig:manifold_demo} to help demonstrate some of the assumptions:
\begin{enumerate}[label=\textbf{A.\arabic*}]
    \item \label{A1} The given data is a local subset of the smooth data manifold $\cM\cap U$ with $U\subset\bbR^d$ centered at $\v x_0$.
    \item \label{A2} The linear regression problem is investigated under a local coordinate frame of $\bbR^d$, where $\v x_0 := \v 0$ and the tangent space $\cT_{\v x_0}\cM$ at $\v x_0$ is the independent variable space parameterizing the smooth data manifold locally.
    \item \label{A3} The data is uniformly distributed in a subset $\Omega'$  of the independent variable space $\cT_{\v x_0}\cM$, \emph{i.e.}, $\rho(\v x) = \frac{1}{\abs{\Omega'}}$:\begin{equation}\label{eqn:uniform}
    \expt{\,\boldsymbol\cdot\,} = \frac{1}{\abs{\Omega'}}\int_{\Omega'}\;\boldsymbol\cdot\; \dd{\v x}.
        \end{equation}
    \item \label{A4} All the independent variables $x_i$ are assumed to be $i.i.d.$ according to $\cU\big([-L,L]\big)$ where $L \ll 1$ determines the size of the local region $\Omega'$.
\end{enumerate}
\end{assumptions}

While a more general assumption could be entertained for the data distribution, we contend that the above
formulation provides more clarity in the derivations of the linear regression solutions and explicates the primary influence exerted by the local geometry of the underlying manifold.

\subsection{{Simple curves in $\mathbb{R}^2$}}\label{sec:toy_model}
We first consider 
a model problem in which $\cM$ is given as the curve {$\cM := \big(x, y(x)=\kappa x^2\big)$} in $\mathbb{R}^2$ with $\kappa$ characterizes the curvature of $\cM$ at $(0,0)$. {Since the tangent space at $(0,0)$ is already given as the $x$-axis, \ref{A2} is automatically satisfied. According to \ref{A3}-\ref{A4}, the local data manifold is given by $\big(x, y(x)\big)$ where $x\in [-L, L]$ and $\rho(x)=1/2L$.}
Then, \cref{eqn:lin_systm} for this model problem becomes the following $3\times 3$ system:  
\begin{equation} \label{eqn:toy_system}
    \begin{bmatrix}
\left <x^2\right > & \left <xy\right > & \left <x\right > \\
\left <xy\right > & \left <y^2\right > & \left <y\right > \\
\left <x\right > & \left <y\right > & 1
\end{bmatrix}
\begin{bmatrix}
w_x\\
w_y\\ 
b
\end{bmatrix}
 = 
 \begin{bmatrix}
\left <gx\right >\\
\left <gy\right >\\ 
\left <g\right >
\end{bmatrix}.
\end{equation}
{With \ref{A3}-\ref{A4}}, we have $\left <xy\right >=\left <x\right >=0$, and \cref{eqn:lin_systm} is further reduced to 
$$
\begin{bmatrix}
\left <x^2\right > & 0 & 0 \\
0 & \left <y^2\right > & \left <y\right > \\
0 & \left <y\right > & 1
\end{bmatrix}
\begin{bmatrix}
w_x\\
w_y\\ 
b
\end{bmatrix}
 = 
 \begin{bmatrix}
\left <gx\right >\\
\left <gy\right >\\ 
\left <g\right >
\end{bmatrix}. 
$$
The solution is given by 
$$
\begin{dcases}
w_x^* = \frac{\left <gx\right >}{\left <x^2\right >},\\
w_y^* = \frac{1}D(\left <gy\right >-\left <g\right >\left <y\right >),\\
b^* = \frac{1}D(-\left <y\right >\left <gy\right > + \left <y^2\right >\left <g\right >),
\end{dcases}
$$
where $D$ is the determinant $D := \left <y^2\right >-(\left <y\right >)^2$.  If $g$ is $C^2$:
\begin{align*}
    g\big(x,\,y(x)\big) = g(0) + \pp{g}{x}(0)x +\pp{g}{y}(0) y + \frac{1}{2}\big(\pp{^2g}{x^2}(0)x^2 + \pp{^2g}{xy}(0)xy + \pp{^2g}{y^2}(0)y^2\big)+ O(|x|^3). 
\end{align*}
 Combining with $\left <xy\right >=\left <x\right >=0$, we finally have 
$$
\begin{dcases} \label{eqn:2d_sol}
w_x^* = \pp{g}{x}(0) +\cO\big( L^2\big), \\
w_y^* = \pp{g}{y}(0) + \frac{1}{2\kappa}\pp{^2g}{x^2}(0)  +\mathcal{O}\big(L^2\big),\\
b^* = g(0)+\mathcal{O}\big(L^4\big),
\end{dcases}
$$
{where the power of $L$ terms are considered to be of higher order since we take $L\ll 1$}. 

\begin{figure}[ht]
\vskip 0.2in
\begin{center}
\centerline{\includegraphics[width=0.4\textwidth]{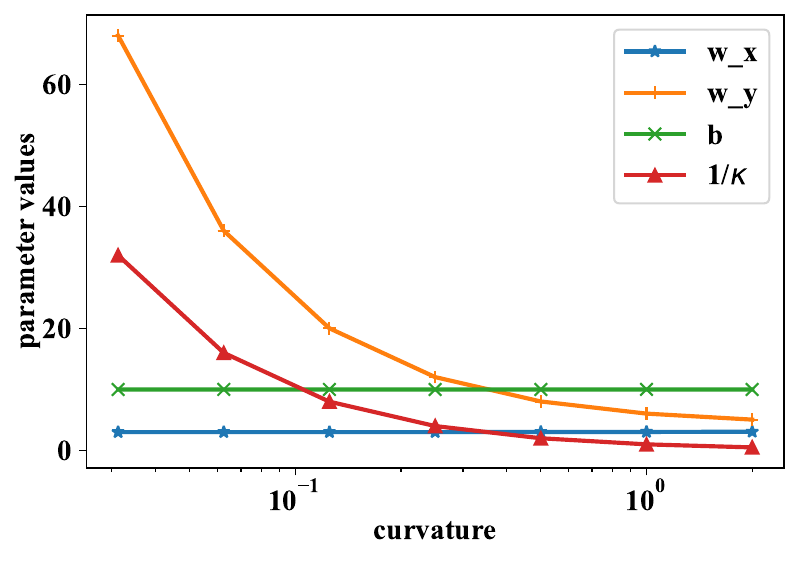}}
\caption{$2$D linear regression solutions on data manifolds $(x, \kappa x^2)$ with different curvatures $\kappa$. $x$-axis is in log-scale. When curvature is small, the optimal solution $w_y^*$ is dominated by the curvature effect, in a magnitude related to $\frac{1}{\kappa}$ as we have shown in~\eqref{eqn:2d_sol}.}
\label{fig:toy_blowup}
\end{center}
\vskip -0.2in
\end{figure}

Based on the derived formulas, we observe that the magnitude of $w_y^*$ tends to blow up when the curvature of the underlying curve $\cM$ goes to $0$. To demonstrate this claim, we perform a simple numerical simulation on the linear regression problem with the target function  $g(x,y) = 2x^2 + 2y^2 + 6xy + 3x + 4y + 10$. We randomly sample $N=1000$ points uniformly in the interval $[-0.1, 0.1]$ to obtain $x$, and apply the linear regression solver to fit the data function $g(x,y)$, for different values of the curvature $\kappa$. The simulated result shown in \Cref{fig:toy_blowup} solidifies our assertion: $w_y$ tends to blow up as the curvature approaches $0$.

To summarize, we list the following properties which will persist in more general situations discussed in this paper: 
\begin{enumerate}
    \item \textbf{The well-posedness of the linear regression.}\\ The matrix $\begin{bmatrix}
    \left <y^2\right > & \left <y\right > \\
 \left <y\right > & 1
\end{bmatrix}$ will not be invertible if $\left <y^2\right >=0$ due to $\kappa= 0$. The data manifold is therefore called ``flat" at the origin. 
    \item \textbf{The effect of geometry on the solutions}\\  
    In the case of non-zero curvature, the linear regression problem has a unique solution. However, one must account for the influence of the curvature when attempting to approximate the first-order information of $g$ (unless $g$ is linear). This is particularly relevant \emph{when the curvature is small as it could lead to contamination of the target solution}. Therefore, the impact of geometry on the solutions in a more general setup should not be overlooked.

\end{enumerate}
 
 


\subsection{Hypersurfaces in $\bbR^d$} \label{sec:hypersurface}

In this section, we derive results similar to those obtained in \Cref{sec:toy_model}, which allow us to address the aforementioned inquiries for general hypersurfaces.
The hypersurface $\cM$ in $\bbR^d$ is an embedded submanifold with codimension $1$. Assume $\cM$ is locally smooth ($C^m$ for any $m\ge 1$). Then by the inverse function theorem, $\cM$ can be locally represented by the graph of a unique continuously differentiable function, meaning there is a smooth mapping $h$ (of full rank) of $(d-1)$ variables defined in the $(d-1)$-dimensional neighborhood $U'$ of $\v x_0$ $s.t.$
\begin{equation} \label{def:graph_rep}
	\cM\cap U = \big\{ x_d =y = h(\v x'): \v x' \in U'\subset\bbR^{d-1} \big\},
\end{equation}where $\v x \in\bbR^d= (x_1, \, x_2, \, \cdots,\,x_d) = (\v x',\, x_d)$, and $U$ is a neighborhood of $\v x_0\in \bbR^d$.

Therefore, the tangent space $\cT_{\v x_0}\cM$ at $\v x_0$ is identified by a hyperplane in $\bbR^d$ implying the normal space is only $1$-dimensional since by definition the normal space is spanned by all the unit vectors $\v n(\v x_0)$ at $\v x_0$ that satisfy $\inpd{\v n}{\v v}=0$ for all $\v v \in \cT_{\v x_0}\cM$, where $\inpd{\cdot}{\cdot}$ is the standard inner product for the embedding space $\bbR^d$. To extend the concept of curvatures from \Cref{sec:toy_model}, we can rely on the shape operator~\cite{singer2015lecture} from differential geometry:
\begin{proposition}\label{thm:eigenvalue}
The principal curvatures $\kappa_1, \kappa_2, \dots, \kappa_{d-1}$ (associated with $\v n(\v x)$) of $\cM$ at $\v x_0$ defined through the shape operator are given by the eigenvalues of the Hessian at $\v x_0$: $Hess(h)(\v x_0)$ of the graph representation $h$. 
\end{proposition}

{\Cref{thm:eigenvalue} establishes a direct relationship between the extrinsic principal curvatures and the Hessian, the generalized second-order derivative, of the graph of $\cM$. The following Lemma further motivates and generalizes the surface representation assumed in \Cref{sec:toy_model}. }

\begin{lemma}\label{thm:hypersurface_represent}(Local representation of the hypersurface $\cM$)\\
Locally under a suitable coordinate basis, the hypersurface $\cM$ can be approximated by a quadratic form using the principal curvatures up to the second order.
\end{lemma}
\begin{proof}
    The local graph representation of $\cM$ is given by $y = h(\v x')$ for $\v x'\in U'\subset\bbR^{d-1}$, a neighborhood of $\v x'_0$. Without loss of generality, we take $\v x'_0 = \v 0_{d-1}$. A Taylor expansion of $h$ around $\v 0_{d-1}$ gives:
    \begin{align*}
        y = h(\v x') = h(\v 0_{d-1}) + \nabla h(\v 0_{d-1})\cdot\v x'+ \frac{1}{2}\v x'^THess\big(h(\v 0_{d-1})\big)\v x' + \cO(\norm{\v x'}^3).
    \end{align*}
    Since $\v x = \v 0 \implies h(\v 0_{d-1}) = 0$, and $ \nabla h(\v 0_{d-1})=\v 0_{d-1}$ because $h$ is tangent to the $x$-plane, we have $y = h(\v x') = \frac{1}{2}\v x'^THess\big(h(\v 0_{d-1})\big)\v x' + \cO(\norm{\v x'}^3)$. The Hessian is semi-positive definite, by diagonalization and a change of basis, we eventually have 
    \[y=h(\v x') = \frac{1}{2}\v x'^T\cK\v x' + \cO(\norm{\v x'}^3),\]
    where $\cK$ is a diagonal matrix with its diagonal being the principal curvatures $\kappa_1, \kappa_2, \dots, \kappa_{d-1}$ based on \Cref{thm:eigenvalue}. Thus $\cM$ can be approximated up to the second order by:
    \(y \approx \sum_{i=1}^{d-1} \kappa_ix_i^2.\)
\end{proof} With the above local approximation, we can obtain a local linear regression formula for hypersurface around $\v x_0$:
\begin{theorem}(Solution formulas for local linear regression on hypersurfaces)\label{thm:hypersurface}\\
When the data manifold $\cM$ is a hypersurface with a local approximation $(\v x', y=\sum_{i=1}^{d-1} \kappa_ix_i^2)$ given in \Cref{thm:hypersurface_represent}, under the assumptions \ref{A2}-\ref{A4}, if the linear regression problem is well-posed, it has the following solution:
\[
\begin{dcases}
 w_i* = \pp{g}{x_i}(\v 0) + \cO(L^2),\\
	w_y^* = \pp{g}{y}(\v 0) + \frac{1}{2}\frac{\displaystyle\sum_{i=1}^{d-1}\kappa_i \pp{^2g}{x_i^2}(\v 0)}{\displaystyle\sum_{i=1}^{d-1}\kappa_i^2} + \cO(L^2), \\
	b^* = g(\v 0) + \cO \Bigg(\sum_{i=1}^{d-1} \pp{^2g}{x_i^2}(\v 0)\Big(\frac{\displaystyle\sum_{j\neq i}\kappa_j^2 - \kappa_i\sum_{j\neq i}\kappa_j}{\displaystyle\sum_{k=1}^{d-1}\kappa_k^2}\Big)\frac{L^2}{3} \Bigg).
\end{dcases}\]
\end{theorem}
\begin{proof}
Following a similar argument and the same notation introduced in \Cref{sec:toy_model}, by symmetry the linear system resulted from the least square minimization is given by:
\begin{equation}\label{eqn:lin_systm_nd}
	\begin{bmatrix}
		\expt{x_1^2} & 0 &  & \dots & 0 &0\\
		0 & \expt{x_2^2} & 0 & \dots &0 & 0\\
		\vdots &   \ddots & \ddots & \ddots & \vdots & \vdots\\
		0 & 0 &  \dots &0 & \expt{y^2}& \expt{y}\\
		0 & 0& \dots & 0 & \expt{y}& 1
	\end{bmatrix}\begin{bmatrix}
	w_1 \\ w_2 \\ \vdots \\ w_{d-1} \\ w_{y}\\ b
\end{bmatrix}= \begin{bmatrix}
	\expt{gx_1} \\ \expt{gx_2} \\ \vdots \\\expt{gx_{d-1}}  \\ \expt{gy} \\ \expt{g}
\end{bmatrix}.
\end{equation}
Then it is easy to see $w_i$ is trivial to obtain, and we again only have a $2\times 2$ system to solve for $w_y$ and $b$, similar as in \Cref{sec:toy_model}. The rest follows from some algebraic manipulations, see \Cref{sec:hypersurface_proof} for details.
\end{proof}

To conclude for the local linear regression on hypersurfaces, we note that we obtain a generalized system with similar structures as in the case of \Cref{sec:toy_model}, however, the previous problematic situations become much nicer:
\begin{enumerate}
    \item The system is invertible as long as we have at least one $\kappa_i\neq 0$. When $\kappa_i=0$, corresponding higher-order nonlinearity assumes the role of $\kappa_i$,  meaning the hypersurface only needs to be non-flat in at least one direction.
    \item The blow-up effect in $w_y^*$ caused by curvatures is mitigated: as long as $k_i$ is large in some directions, the term  $\frac{1}{2}\displaystyle\sum_{i=1}^{d-1}\kappa_i \pp{^2g}{x_i^2}(\v 0) / \displaystyle\sum_{i=1}^{d-1}\kappa_i^2$ would be well-controlled.
\end{enumerate}

\subsection{Codimension-$k$ submanifolds} \label{sec:codim-k}

The success of obtaining a solution formula in \Cref{sec:hypersurface} is mainly due to the reduction to a $2\times 2$ system in the end of the least square minimization, a direct consequence of the fact that the normal space is always $1$-dimensional for a hypersurface. In general, for a codimension-$k$ submanifold whose normal space is therefore $k$-dimensional, one has to solve a $(k+1)\times (k+1)$ linear system from the least square minimization under the same assumption, hence the solution formula cannot be easily obtained.

Furthermore, from \Cref{thm:eigenvalue} we learn that the diagonalization process and its connection to principal curvatures are strongly predicated on the assumption of maintaining a consistent normal direction. For a $k$-dimensional normal space, there will be one shape operator associated with each orthonormal basis of the normal space, meaning whenever we diagonalize the Hessian of the graph representation along one normal direction and obtain a change of coordinates for the tangent space, the resulting local coordinate frame generally would not diagonalize the others.

To solidify the above statements, consider the example of a $2$-manifold in $\bbR^4$, which is a codimension-$2$ submanifold with $2$ independent normal basis vectors. Following the same procedure as in \Cref{sec:hypersurface}, we would obtain locally a quadratic graph representation $\big (x_1, x_2, y_1 = k_{11}x_1^2 + k_{22}x_2^2\big )$ when restricted on the linear subspace $\cT_{\v x_0} \cM \oplus \cN_1$. With this coordinate frame, locally around $\v x_0$, $\cM$, in general, would take the following form up to some higher order errors:
\begin{align*}
    \cM_{\v x_0} = \big \{(x_1, x_2, y_1, y_2)\, \big \vert\, y_1 = k_{11}x_1^2 + k_{22}x_2^2 + \dots, y_2 = \tau_{11}x_1^2 + \tau_{12}x_1x_2 + \tau_{22}x_2^2 +\dots\big\}, 
\end{align*} 
where the graph representation along $y_2$ is not diagonalized. Besides, after applying local linear regression as in \Cref{sec:hypersurface}, the resulting linear system from the least square minimization is given as:
\begin{equation*}
	\begin{bmatrix}
		\expt{x_1^2} & 0 & 0 & 0 & 0\\
		0&  \expt{x_2^2} &0& 0 & 0\\
		0& 0 & \expt{y_1^2}&  \expt{y_1y_2} & \expt{y_1}\\
		0& 0 & \expt{y_1y_2} &\expt{y_2^2}   & \expt{y_2}\\
		0& 0 & \expt{y_1} & \expt{y_2}& 1
	\end{bmatrix}\begin{bmatrix}
		w_{x_1} \\ w_{x_2} \\ w_{y_1} \\ w_{y_2} \\ b
	\end{bmatrix}= \begin{bmatrix}
		\expt{gx_1} \\ \expt{gx_2} \\ \expt{gy_1} \\ \expt{gy_2} \\ \expt{g}
	\end{bmatrix},
\end{equation*}
implying to obtain the solution formulas, one needs to solve the $3\times 3$ dense linear system related to $y_1,\,y_2$, and $b$, 
\[\begin{bmatrix}
		\expt{y_1^2}&  \expt{y_1y_2} & \expt{y_1}\\
		\expt{y_1y_2} &\expt{y_2^2}  & \expt{y_2}\\
		\expt{y_1} & \expt{y_2}& 1
		\end{bmatrix}\begin{bmatrix}
			w_{y_1} \\ w_{y_2} \\ b\end{bmatrix} = \begin{bmatrix}
				\expt{gy_1} \\ \expt{gy_2} \\ \expt{g} \end{bmatrix}, \]
where in general a solution formula is hard to obtain, so is the explicit effect of the data geometry on the solutions. Nonetheless, investigating the well-posedness of the problem remains valuable. Apparently, a sufficient condition for the linear system to be singular, without loss of generality, is when $y_2\equiv 0$, indicating a completely flat projection of $\cM$ onto the linear subspace $\cT_{\v x_0} \cM \oplus \cN_2$. In general, this may occur when a low-dimensional manifold is embedded within a high-dimensional space and lacks sufficient non-linearity for any of the normal directions in the chosen local coordinate frame. Therefore, we can state the following theorem:

\begin{theorem} (Ill-posedness of linear regression for codimension-$k$ submanifolds) \label{thm:codim-k}
For a codimension-$k$ submanifold $\cM$ embedded in $\bbR^d$ where $k\geq 1$, the local linear regression on $U\cap\cM$ for $U\subset\bbR^d$ leads to solving a $(k+1)\times (k+1)$ linear system, which could be ill-posed if the submanifold restricted on $U$: $U\cap\cM$ is flat in any of its normal direction in a chosen local coordinate frame.
\end{theorem}

For those cases identified as problematic in \Cref{thm:codim-k}, \emph{i.e.}, the local graph representation of $\cM$ along some normal direction $y_i$ is $0$, where $\cM$ is given locally by $\mathcal M |_{\v x_0} = \left\{ (x_1, \cdots,x_{d-k}, y_1,\cdots, y_k)\right\}$,
it is important to note that the reach \cite{federer1959curvature}, even for the restricted subset $U\cap\cM$, can still be finite, where the finiteness merely guarantees that the manifold is non-flat in {some directions, not all.} Such findings suggest that the implications derived from the reach measure may not be sufficient in some practical settings. 

A direct comparison between \Cref{thm:codim-k} and \Cref{thm:hypersurface} reveals the distinct contributions of the intrinsic dimension and the codimension of the data manifold. Specifically, as the intrinsic dimension increases, the undesirable behavior in $w_y^*$ is less likely to occur. Conversely, if the data manifold can be considered as a subset of $\bbR^{\gamma}$, care must be taken when isometrically embedding it into $\bbR^d$ with $d\gg \gamma$, as the manifold may not possess adequate non-linearity in all directions of the normal space, thereby leading to an ill-posed linear regression problem. Nonetheless, we demonstrate  in \Cref{sec:noise} the benefits of noise to prevent the degeneracy of the linear regression problem so that this concern rarely arises in practical applications.

Finally, while an explicit solution formula for general codimension-$k$ submanifolds remains elusive, our study demonstrates that a solution formula can still be derived for a specific case involving a codimension-$(d-1)$ submanifold in $\bbR^d$, namely a curve embedded in $\bbR^d$ in \Cref{sec:curve}. The derivation utilizes the concept of the Frenet-Serret frame from classical differential geometry and techniques akin to the method of matched asymptotic expansion.

\subsection{Experiments with MNIST dataset}\label{sec:mnist}
To gain further insights into \Cref{thm:codim-k} and more practical implications of the linear regression problem, we conducted numerical experiments with the MNIST dataset \cite{lecun1998gradient}. 
Specifically, we use the {1032} images related to the digit $2$ as data points in $\bbR^{784}$ (each of these $28\times 28$ pixels are regarded as a point in $\bbR^{784}$). We then construct a linear data function, denoted by $g_{784}$, by generating a random Gaussian vector in $\bbR^{784}$ and normalizing it to have unit norm. This normalized vector $\v u = (u_1, u_2,\dots, u_{784})$ defines the linear function linear $g_{784}$ as $g_{784} = \v u \cdot \v x = u_1x_1 + u_2x_2 + \dots u_{784}x_{784}$.

\begin{figure}[htbp!]
    \centering
      \subfigure[]{\includegraphics[width=0.45\linewidth]{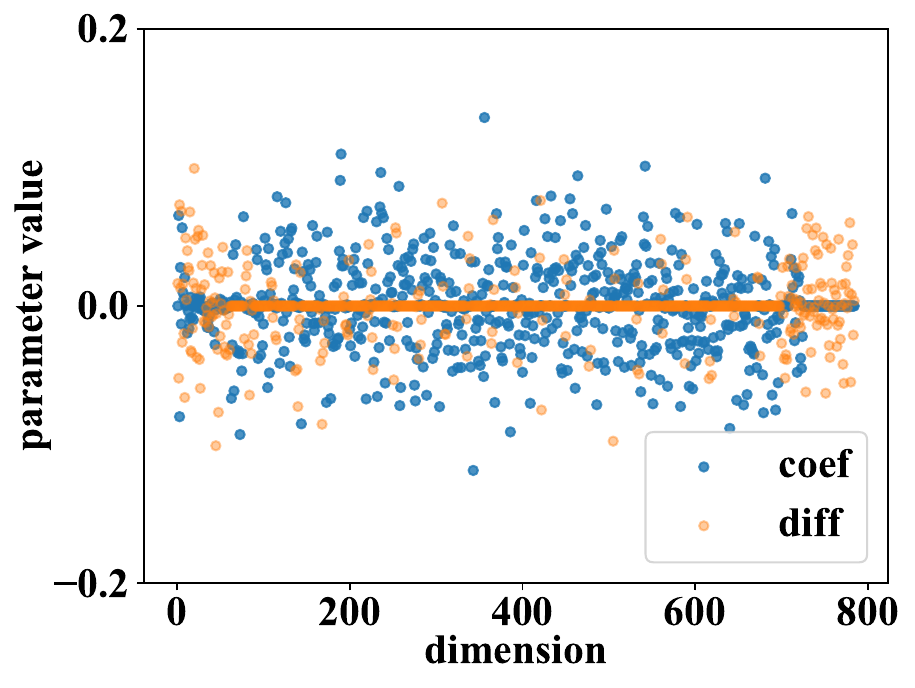}
               \label{fig:reg_full}}
      \hfill
      \subfigure[]{
        \includegraphics[width=0.45\linewidth]{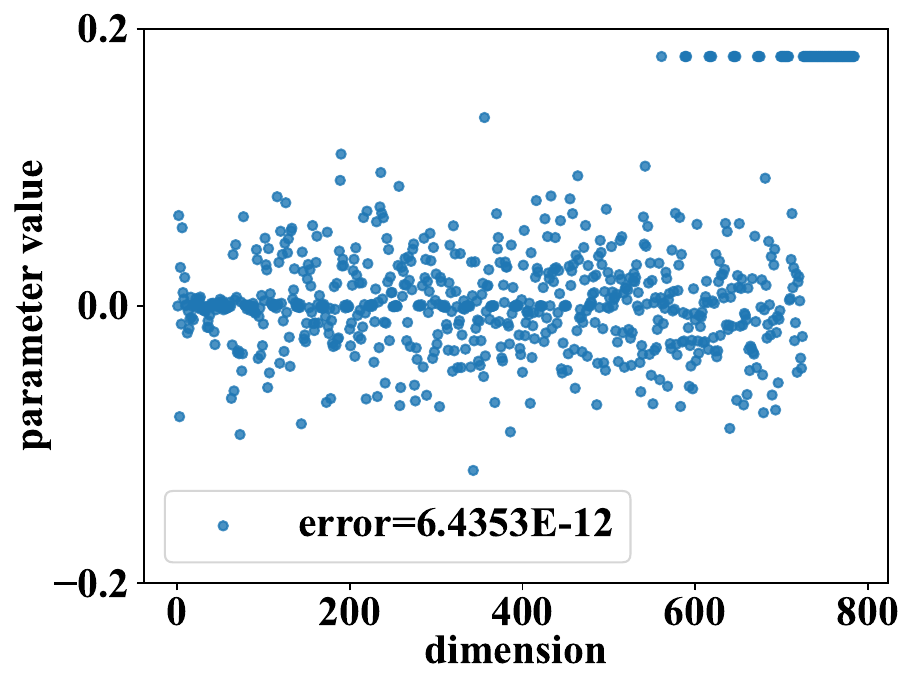}
          \label{fig:reg_full_perturb}}
\caption{ \emph{Left:} blue dots represent computed linear regression solution $\v w^* =(w^*_1,\dots,w^*_{784})$, while orange dots show $\v w^*-\v u$. A constantly-zero tail can be observed at the end for blue dots. \emph{Right:} all parameter values with absolute values  $<10^{-8}$ are changed to $0.18$, the error magnitude stays the same.}
\end{figure}

We employ the standard scikit-learn software \cite{pedregosa2011scikit} to
solve the least square problem $\min_{\v w}|| \v w\cdot \v x - g_{784}||_2^2$.
The resulting solutions $\v w^*$ are shown in \Cref{fig:reg_full}. Notably the obtained solutions differ from the expected solution $\v u$, while the pointwise approximation error is close to machine epsilon: $\norm{g_{784}(\v x)-\tilde{g}_{784}(\v x)} < 6.5\times 10^{-12}$, where $\tilde{g}_{784}$ is the linear regression approximation with magnitude $\sim\cO(10^3)$. Different parameters leading to essentially the same evaluation indicates that 
the linear regression is undetermined.

When the linear system is undetermined, the linear regression software is designed to find the least $L_2$-norm solution to ensure uniqueness, which corresponds to introducing \emph{regularizer} to the original linear regression problem. This regularization causes the free parameters to always be $0$, which forms the constantly-zero tail in \Cref{fig:reg_full}. Furthermore, we artificially modify the parameter values in the obtained solution: we change all the parameter values with an absolute value $<10^{-8}$ to a fixed value of $0.18$ as shown in  \Cref{fig:reg_full_perturb}. These parameter values should correspond to  dimensions that do not contribute to the final evaluation of $\tilde{g}_{784}$. Therefore, arbitrarily perturbing them should not have a substantial impact. Interestingly and as expected, the error in the resulting point-wise evaluation remains the identical as in the previous case, providing evidence that these dimensions are indeed redundant.

\begin{figure}[htbp!]
    \centering
      \subfigure[]{\includegraphics[width=0.45\linewidth]{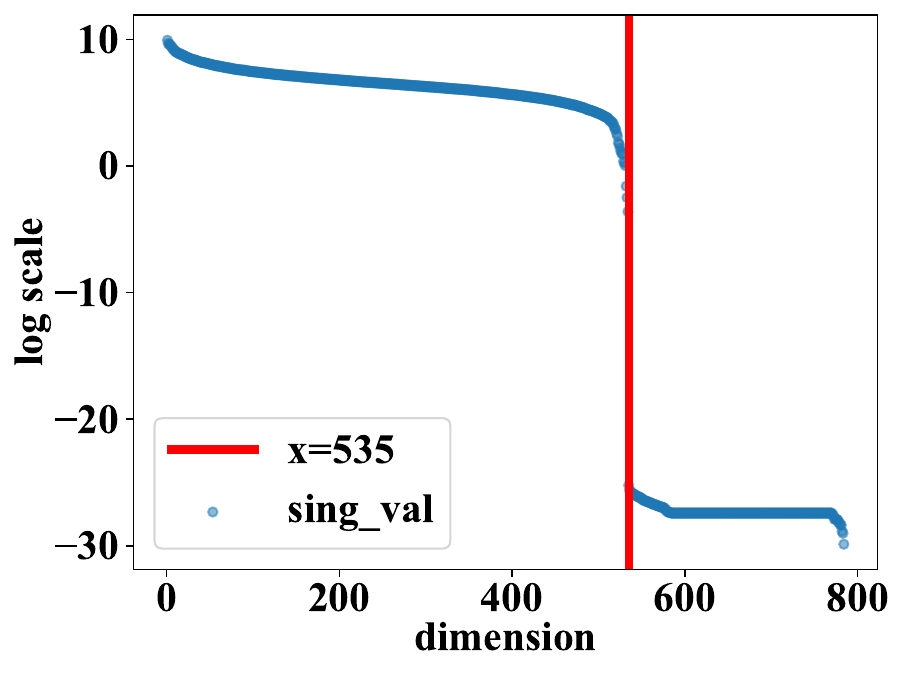}
            \label{fig:sing_val}}
      \hfill
      \subfigure[]{
        \includegraphics[width=0.45\linewidth]{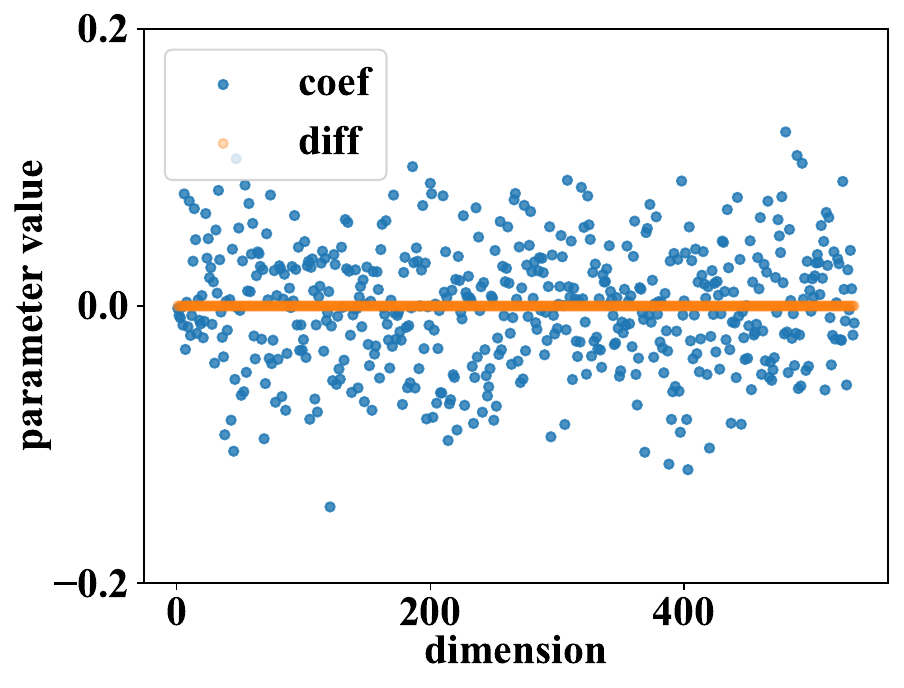}
          \label{fig:reg_534}}
\caption{\emph{Left:} singular value plots of the digit-$2$ images in $\v 0$-centered MNIST under log scale.  A sharp decay is observed at around  $\v q_{535}$.  \emph{Right:} linear regression solution on the space spanned by $\{\v q_1, \dots \v q_{534}\}$.}
\end{figure}

\Cref{thm:codim-k} suggests that the ill-posed nature of the linear regression problem may be the consequence of the submanifold, where the digit-$2$ MNIST images reside, is confined in some lower dimensional linear subspace. Analyzing the singular value of the $\v 0$-centered dataset in \Cref{fig:sing_val}, we note a $534$D linear subspace can effectively capture the dataset. This implies the digit-$2$ MNIST submanifold, embedded in $\bbR^{784}$, exhibits nonlinearity up to around $534$ dimensions, while it is flat with respect to the remaining dimensions. 

To validate this observation, let $\{\v q_1, \v q_2, \dots, \v q_{784}\}$ denote the principal components of the dataset, and we project the digit-$2$ data onto $\bbR^{534}$ using $\{\v q_1, \dots, \v q_{534}\}$, and repeat a $534$-dimensional linear regression experiment similar to the one on $\bbR^{784}$. In \Cref{fig:reg_534}, we observe that the linear regression recovers the parameters with no constantly-zero tail, indicating that the regression solution is unique.
 
We now examine the impact of nonlinearity in the data distribution using data functions of the form: $g(\mathbf{x}) = 0.02\sum^{784}_{i=1} x_i^2 + 0.01\sum^{784}_{i=1} x_i$.  
Let $\tilde{\v x} = (\tilde{x}_1,\dots, \tilde{x}_{784}) \in \bbR^{784}$ denote the projected data on $span\{\v q_1, \dots, \v q_{534}\}$, we introduce random quadratic polynomials along \emph{all} the normal directions $\{\v q_{535},\dots,\v q_{784}\}$ to bend the data, \emph{i.e.}: 
\[ \bar{\v x}(\alpha) = \tilde{\v x} + \alpha\sum^{784}_{i=535} (\sum_{j=1}^{784} \eta_{ij}\tilde{x}_j^2) \v q_i,\] where $\eta_{ij}\sim\cN(0,1)$ and $\alpha$ is the parameter to control the degree of bending. Next, we perform linear regression on $\bar{\v x}(\alpha)$ with varying $\alpha$ to approximate the quadratic function $g$. We record the magnitude of the resulting solutions $\v w$ as well as the magnitude of the approximation error $e_1 = \norm{\v w\cdot\bar{\v x}(\alpha)-g(\bar{\v x})}$ in \Cref{tab:mnist_bend}. We observe that as the degree of bending decreases, the pointwise approximation error on the in-sample data remains unchanged, while the magnitude of the resulting solutions increases significantly, echoing the phenomenon observed in \Cref{sec:hypersurface}. 

Furthermore, we create out-of-sample data points by extending along the combined normal direction: $\v x_{out} = \tilde{\v x} + 0.1\sum_{i=535}^{784} \v q_i$, we found the ``off-sample" evaluation error $e_2 = \norm{\v w\cdot\v x_{out}-g(\v x_{out})}$ also drastically increases as the curvature decreases (column $4$), indicating the small amount of bending of the data manifold adversely affects the generalization to out-of-sample data. 

\begin{table}[!htb]
    \centering
    \caption{Magnitude of different tests of approximating a quadratic function using linear regression on full-dimensional curved data. The bending magnitude affects the linear regression solution magnitudes (column $2$) and the generalization errors (column $2$).}
    \begin{tabular}{r|r|r|r}
        $\alpha$ & $\norm{w}_2$ & $e_1$ & $e_2$ \\
        \hline
        1e-8 & $92.27$ & $463.21$ & $2580.06$ \\
        \hline
        1e-10 & $7678.69$ &  $463.21$ & $10073.15$ \\
        \hline
        1e-12 & $767852.03$ & $463.21$ & $1269868.67$\\
    \end{tabular}
    \label{tab:mnist_bend}
\end{table}

\section{Regularization from noise} \label{sec:noise}

In \Cref{sec:regression_formula_manifold}, we show that local linear regression for data distributed on submanifolds can have two potential issues: (1) the problem may no have a unique solution when the data manifold is flat in one of the normal directions; (2) the nonlinearity of the data manifold could have a nontrivial effect on the first order information of the target function  $g$. In this section, we investigate how the presence of noise can potentially alter these scenarios.

\subsection{Noisy data could prevent degeneracy}
He et al.~\yrcite{he2023side} pointed out that the presence of noise in the ambient space around the data manifold can regularize the linear regression problem. We recount their finding specifically for the simple problem considered in \Cref{sec:toy_model}. In their setting, the data is distributed just on the $x$-axis (effectively by setting the curvature $\kappa =0$,  so along the $y$-axis the data manifold is flat). Under the presence of Gaussian noise in the $y$ component of the data coordinates, \emph{i.e.}, $y=\sigma\eta\sim\cN(0,\sigma^2)$ instead of $y\equiv 0$,  the optimal parameter $w_y^*$ would converge with high probability to:
\[ w_y^* \sim \pp{g}{y} + \cO(\frac{1}{\sigma\sqrt{N}}).\]
This implies that in the presence of noise, the previously degenerated linear system defined by the noise-free data is now invertible, and the true first-order information $ \pp{g}{y}$ is obtained in the limit of the number of data points $N\to\infty$.
This observation can be easily extended to any codimension-$k$ submanifold that is flat in certain directions to address the ill-posed linear regression as discussed in \Cref{sec:codim-k}. A more involved example in $\bbR^4$ is used to demonstrate such point below:

Consider again the case of $2$-manifold embedded in $\bbR^4$ where the graph representation along one normal direction $y_1$ is at least $C^2$, while the graph along the other direction $y_2$ is only linear. Under a suitable local coordinate frame with $y_2 \equiv 0$, the resulting system from the minimization problem is singular if one applies a full-dimensional linear regression. However, with the presence of noise, \emph{i.e.}, $y_2 = \sigma\eta,~~\eta \sim \cN(0,1)$ for some small $\sigma>0$, the linear system under our usual assumption \ref{A1}-\ref{A4} becomes:
\[	\begin{bmatrix}
     x_1 & 0 & 0 &0 & 0\\
     0 & x_2 & 0 &0 & 0 \\
	 0& 0&\expt{y_1^2} & 0 & \expt{y_1}\\
    0  & 0 & 0&\expt{y_2^2} & 0\\
     0&0&\expt{y_1} & 0 & 1
	\end{bmatrix}\begin{bmatrix}
  w_1\\ w_2\\	w_{y_1} \\ w_{y_2} \\  b
\end{bmatrix}= \begin{bmatrix}
	\expt{gx_1}\\ \expt{gx_2} \\ \expt{gy_1} \\\expt{gy_2} \\ \expt{g}
\end{bmatrix}.\]
It is easy to verify that $\expt{y_2^2} = \sigma^2$, leading to $w_{y_2} = \pp{g}{y_2} + \cO(\sigma^2)$. Thus,  the linear system becomes well-posed, and one can obtain the desired first-order information up to the order of {noise variance.}

\subsection{{Interplay between noises and curvatures}}

The second observation concerns the impact of geometry in the derived solution formulas. If one wants to use the local linear regression to obtain first-order information of the function $g$, the geometry of the data manifold would certainly affect the results as derived in \Cref{sec:hypersurface} and \Cref{sec:curve}. Here we show that one simple remedy is to add noises to the data.

We demonstrate the effect of the method by applying it to the case of \Cref{sec:toy_model}. Instead of considering $y = \kappa x^2$, we add the Gaussian noises to $y$ along the $y$-direction to obtain $\tilde{y} = \kappa x^2 + \sigma\eta$. Then, the leading order solution for $w_y^*$ becomes:
\begin{equation*}
     w_y^* \sim \displaystyle\pp{g}{y}(0) + \frac{1}{2}\frac{\kappa L^4}{\kappa^2L^4+\frac{45}{4}\sigma^2}\pp{^2g}{x^2}(0),
\end{equation*}
instead of the previous $w_y^* \sim \pp{g}{y}(0) + \displaystyle\frac{1}{2\kappa}\pp{^2g}{x^2}(0)$. From the derived formula, we note that if we take $\sigma=0$, we recover the formula for the noise-free case. Furthermore, if we take $\sigma^2$ large relative to $\kappa^2L^4$, then the denominator is dominant by $\sigma^2$ such that $\kappa L^4+\frac{45}{4}\sigma^2 \sim \frac{45}{4}\sigma^2$, and the overall term would be small. This suggests that by adding noises of a certain magnitude, the curvature effect would be mitigated and one can obtain a better approximation to the true first-order solution from the linear regression.

\begin{figure}[!ht]
\vskip 0.2in
\begin{center}
\centerline{\includegraphics[width=0.4\columnwidth]{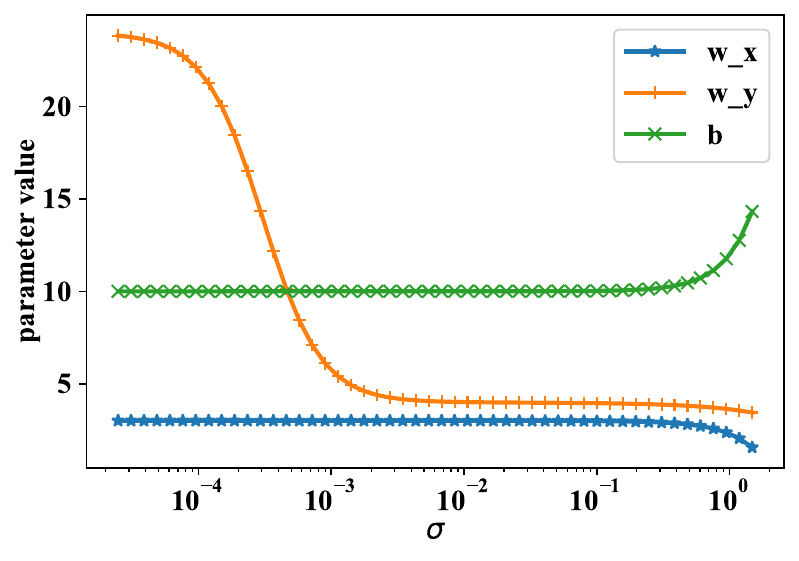}}
\caption{$2$D linear regression solutions under the presence of noises with different $\sigma$. When $\sigma$ is too small, $w_y^*$ still blows up; when $\sigma$ is of certain magnitude, $w_y^*$ approximates $\pp{g}{y}$; but when $\sigma$ is too large, the noise eventually affects $w_x^*$ and $b^*$.}
\label{fig:noise_regularize}
\end{center}
\vskip -0.2in
\end{figure}

To validate the aforementioned observations, we again conduct numerical simulations under the same setting as in \Cref{sec:toy_model} but with added noises. We take $L=0.1$ and $\kappa=0.1$, and vary the noise standard deviation, $\sigma$, to obtain various linear regression solutions; see \Cref{fig:noise_regularize}.

\begin{figure}[!htb]
\centering
\subfigure[]{\includegraphics[width=0.29\columnwidth]{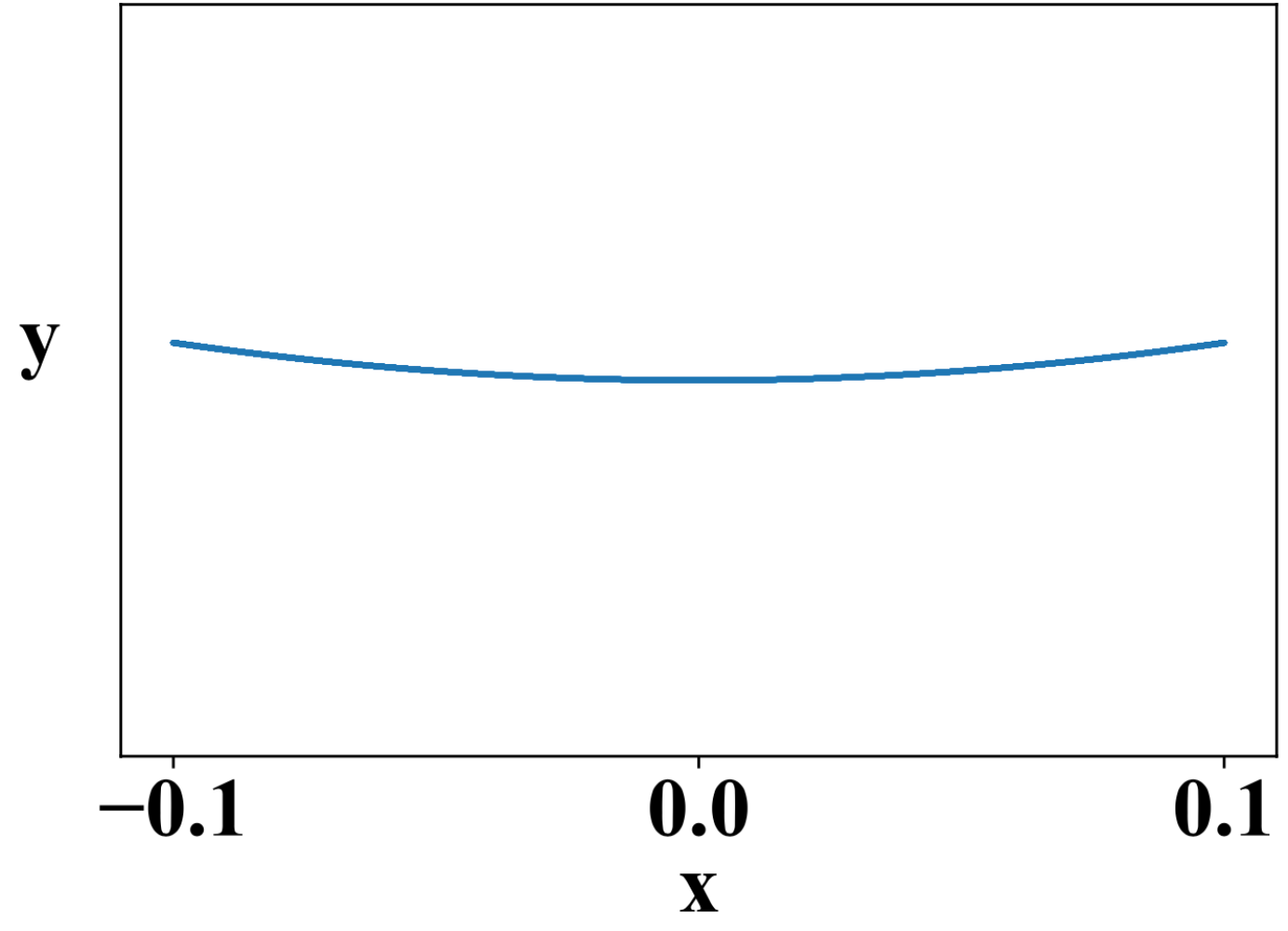}
}
\subfigure[]{\includegraphics[width=0.29\columnwidth]{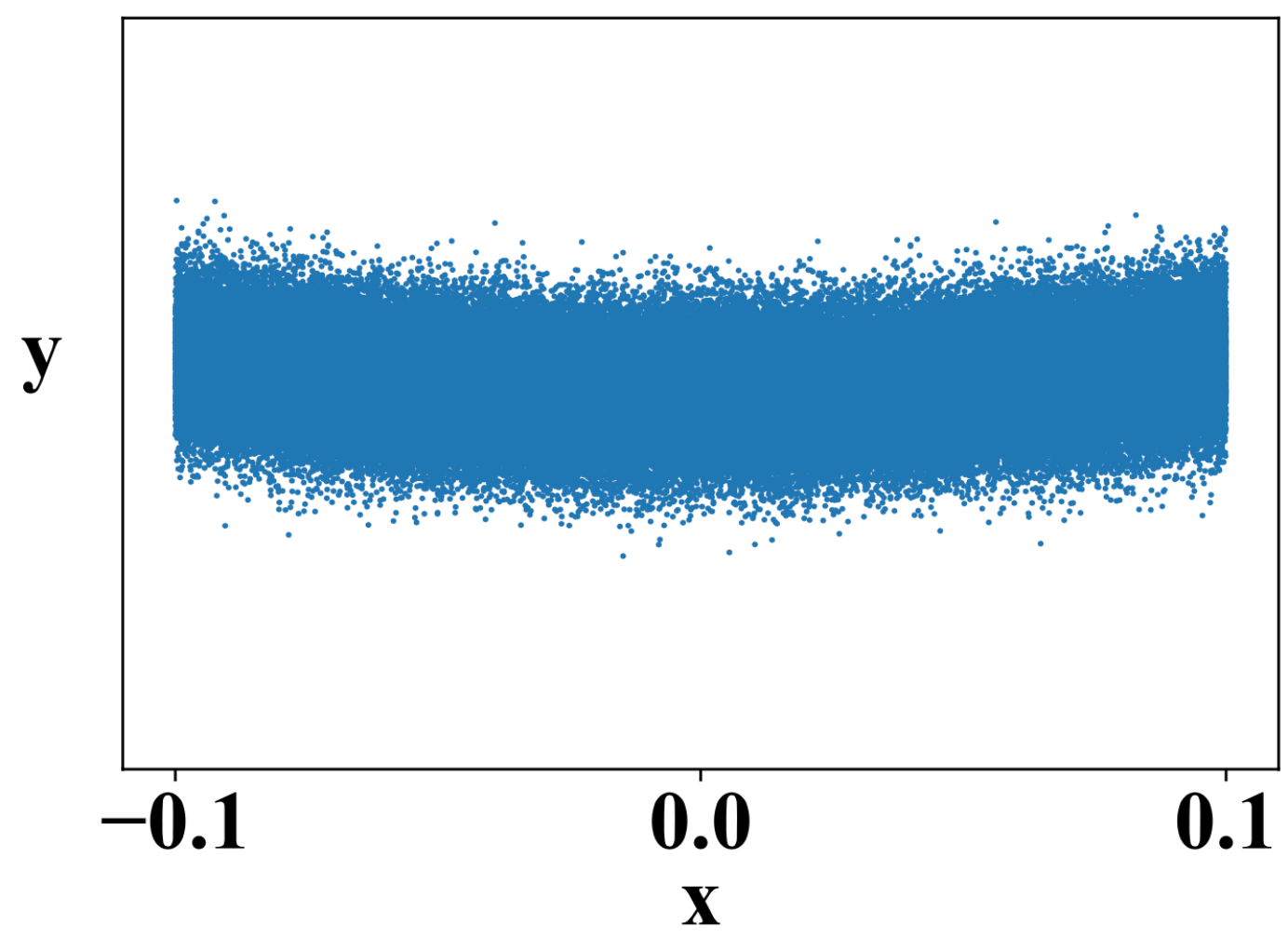}
  }
\subfigure[]{\includegraphics[width=0.29\columnwidth]{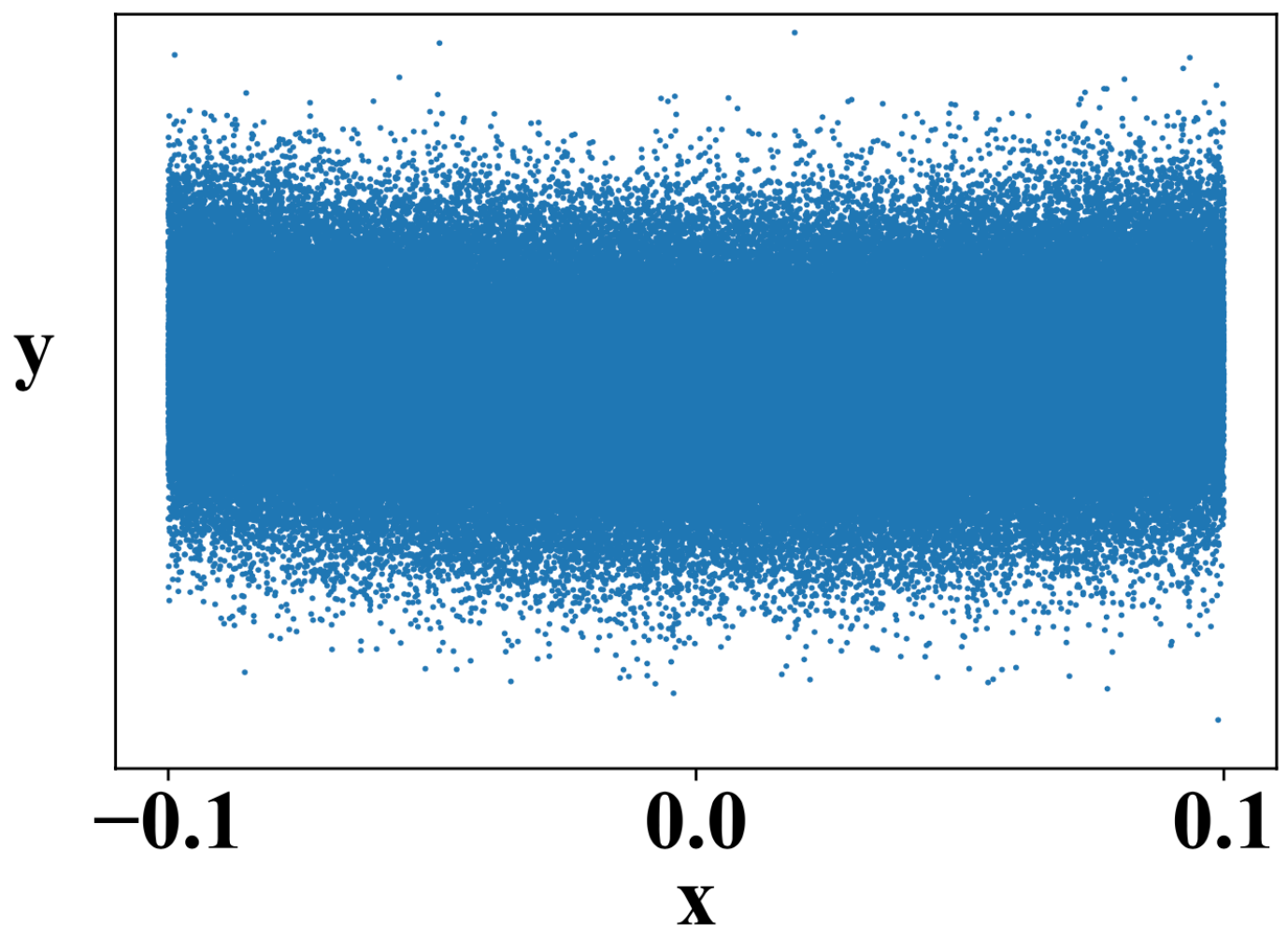}
  }
  \caption{Visualization of the distribution of the noisy data. \\ (a) $\sigma=0, w_y^* = 24$: the solution is influenced by the small curvature, substantially deviates from the target value $\pp{g}{y}=4$;\\ (b) $\sigma=1e-3, w_y^*=5.63$, (c) $\sigma=2e-3, w_y^*=4.43$: when noise is added at a certain level of the standard deviation $\sigma$, $w_y^*$ is significantly reduced; as $\sigma$ increases further, $w_y$ approaches to $4$.}
   \label{fig:noise_vis}
\end{figure}

Remarkably, within a suitable range of noise variance, the blow-up effect arising from a small curvature is mitigated. The underlying mechanism can be better comprehended through the visualizations of the data distribution provided in \Cref{fig:noise_vis}. The intuition behind this is the addition of noise causes the data distribution to blur the structure of the problematic manifold, resulting in an improved approximation to the desired value $\pp{g}{y}(0)=4$.


\section{Conclusions and future work}

This study aims to investigate the impact of data manifold geometry and noise on the well-posedness and stability of local linear regression models for out-of-distribution inferences, both qualitatively and quantitatively. It was found that for general low-dimensional smooth manifolds, the uniqueness of solutions in the local linear regression problem can be compromised when the data manifold is flat in one of the normal directions. Additionally, through theoretical and experimental analysis on specific submanifolds, it was revealed that the nonlinearity of the data manifold, {such as} curvatures, has a significant and nontrivial effect on the stability of the regression outcomes. Furthermore, the presence of noise in the data was shown to not only prevent degeneracy but also interact with curvatures to prevent blow-up in the linear regression solutions.

This work presents a novel approach for analyzing the influence of data manifold geometry and noise on the well-posedness and stability of regression problems. It offers opportunities to incorporate established concepts and techniques from geometry to study diverse data manifolds and unveil their intrinsic impact on regression problems. Additionally, it opens avenues for exploring more complex and practical machine learning scenarios, such as investigating ReLU deep neural networks for regression and convolutional neural networks for classification tasks.

\section*{Acknowledgements}

{We thank all the reviewers for all the constructive suggestions and helpful feedback.} Liu's and Tsai's research is supported partially by National Science Foundation Grants DMS-2110895, DMS-2208504, and by Army Research Office, under Cooperative Agreement Number
W911NF-19-2-0333.





\bibliography{reference}
\bibliographystyle{icml2023}

\newpage
\appendix
\onecolumn
\section{Derivation of solutions for hypersurfaces}\label{sec:hypersurface_proof}

To compute the quantities in each entry in \Cref{eqn:lin_systm_nd}, under the assumption that each $x_i\sim \cU\big([-L, L]\big)$ we have:
\begin{align*}
    \expt{x^2_i}  &= \frac{1}{2L}\int_{-L}^L x^2_i \dd{x_i} = \frac{L^2}3, \\
\expt{y} &= \frac{1}{(2L)^{d-1}}\int_{\Omega'} \sum_{i=1}^{d-1} \kappa_ix_i^2\; \dd{x_1}\dd{x_2}\dots\dd{x_{d-1}} = \frac{1}{(2L)}\sum_{i=1}^{d-1}\int^{L}_{-L} \kappa_ix_i^2\dd{x_i} = \frac{L^2}3\sum_{i=1}^{d-1}\kappa_i,\\
\expt{x_j^2y} &=\frac{1}{(2L)}\int^{L}_{-L} \kappa_jx_j^2\dd{x_j}+ \frac{1}{(4L^2)}\sum_{i=1,\,i\neq j}^{d-1}\int^{L}_{-L} \kappa_ix_i^2 \dd{x_i}\int^{L}_{-L}x_j^2\dd{x_j} = \frac{L^4}5\kappa_j + \sum_{i=1,\,i\neq j}^{d-1}\kappa_i\frac{L^4}{9},\\
\expt{y^2} &= \exptbig{(\sum_{i=1}^{d-1} \kappa_ix_i^2)^2} =\exptbig{\sum^{d-1}_{i=1} \kappa_i^2x_i^4 + 2\sum_{i=2}^{d-1}\sum_{j=1}^{i}\kappa_i\kappa_jx_i^2x_j^2} = \sum_{i=1}^{d-1}\kappa_i^2\frac{L^4}{5} + 2\sum_{i=2}^{d-1}\sum_{j=1}^{i}\kappa_i\kappa_j\frac{L^4}{9}.
\end{align*}
The determinant of the target $2\times 2$ linear system is then given by: \[ D = \expt{y^2} - \expt{y}^2 =  \sum_{i=1}^{d-1}\kappa_i^2\frac{L^4}{5} + 2\sum_{i=2}^{d-1}\sum_{j=1}^{i}\kappa_i\kappa_j\frac{L^4}{9} - \frac{L^4}{9}\big(\sum_{i=1}^{d-1}\kappa_i\big)^2\]
where $\big(\sum_{i=1}^{d-1}\kappa_i\big)^2 = \big(\sum_{i=1}^{d-1}\kappa_i^2 + 2\sum_{i=2}^{d-1}\sum_{j=1}^{i}\kappa_i\kappa_j \big)$. As a result: \[ D = \sum_{i=1}^{d-1}\kappa_i^2(\frac{L^4}{5} - \frac{L^4}9) = \sum_{i=1}^{d-1}\kappa_i^2\frac{4L^4}{45}\]
Then, to obtain an explicit expression for the RHS, we again Taylor expand the data function $g$, assumed to be locally smooth, around the base point: 
\begin{align*}
	&g(\v x) = g\big(\v x',\,h(\v x')\big) = g(\v 0) + \v x^T\nabla g(\v 0)+ \frac{1}2\v x^THess(g)(\v 0)\v x + O(\norm{\v x}^3) \\ 
	=& g(\v 0) + \pp{g}{y}(\v 0) y+ \sum_{i=1}^{d-1} \pp{g}{x_i}(\v 0)x_i +\frac{1}{2}\sum^{d-1}_{i=1} \pp{^2g}{x_i^2}(\v 0)x_i^2  + \sum_{j=2}^{d-1}\sum_{i=1}^{j}\pp{^2g}{x_ix_j}(\v 0) x_ix_j  + \frac{1}{2}\pp{^2g}{y^2}(\v 0) y^2 + O(\norm{\v x}^3)
\end{align*}
where $y^2$ is in fact of $\cO(\norm{x}^4)$. With such expansion along with the previously computed quantities and keeping only the nonzero terms, we can obtain $\expt{g},\, \expt{gx_i}, \expt{gy}$ accordingly:
\begin{align*}
	\expt{g} =& g(\v 0) + \pp{g}{y}(\v 0) \expt{y} + \frac{1}{2}\sum^{d-1}_{i=1} \pp{^2g}{x_i^2}(\v 0)\expt{x_i^2} + \exptbig{\cO(\norm{\v x}^3)}\\
	=&  g(\v 0) + \pp{g}{y}(\v 0)\frac{L^2}3\sum_{i=1}^{d-1}\kappa_i + \frac{1}{2}\frac{L^2}3\sum^{d-1}_{i=1} \pp{^2g}{x_i^2}(\v 0) + \cO(L^4)\\
	\expt{gx_i} =& \pp{g}{x_i}(\v 0)\frac{L^2}{3} + \cO(L^4)\\
	\expt{gy} =& g(\v 0)\expt{y} + \pp{g}{y}(\v 0)\expt{y^2} + \frac{1}{2}\sum_{i=1}^{d-1}\pp{^2g}{x_i^2}\expt{x_i^2y} + \pp{^2g}{y^2}\expt{y^3} + \exptbig{\cO(\norm{x}^6)}\\
	=&g(\v 0)\frac{L^2}3\sum_{i=1}^{d-1}\kappa_i + \pp{g}{y}(\v 0)\big(\sum_{i=1}^{d-1}\kappa_i^2\frac{L^4}{5} + 2\sum_{i=2}^{d-1}\sum_{j=1}^{i}\kappa_i\kappa_j\frac{L^4}{9}\big)\\
	&+\frac{1}{2}\sum_{i=1}^{d-1}\pp{^2g}{x_i^2}(\v 0)\expt{\kappa_ix_i^4 + \sum_{j\neq i}\kappa_jx_i^2x_j^2} + \cO(L^6)\\
	=&g(\v 0)\frac{L^2}3\sum_{i=1}^{d-1}\kappa_i + \pp{g}{y}(\v 0)\big(\sum_{i=1}^{d-1}\kappa_i^2\frac{L^4}{5} + 2\sum_{i=2}^{d-1}\sum_{j=1}^{i}\kappa_i\kappa_j\frac{L^4}{9}\big)\\
	&+\frac{1}{2}\sum_{i=1}^{d-1}\pp{^2g}{x_i^2}(\v 0)\big(\kappa_i\frac{L^4}{5} + \sum_{j\neq i}\kappa_j\frac{L^4}{9}\big)+\cO(L^6)
\end{align*}
Again, the optimal solutions for $w_y,\,b$ are given by applying the inverse (the matrix is non-singular) to the RHS vector $[\expt{gy},\,\expt{g}]^T$, which yields:
$$
\begin{dcases}
w_y^* = \frac{1}D\big(\left <gy\right >-\left <g\right >\left <y\right >\big)\\
b^* = \frac{1}D\big(\left <y^2\right >\left <g\right >-\left <y\right >\left <gy\right >\big)
\end{dcases}
$$
\begin{align*}\label{eqn:wy_nd}
	w_y^* & = \frac{1}{\expt{y^2} - \expt{y}^2}\bigg(\Big(g(\v 0)\expt{y} - g(\v 0)\expt{y}\Big) +  \Big(\pp{g}{y}(\v 0)\expt{y^2} -\pp{g}{y}(\v 0)  \expt{y}^2\Big) \\ & \quad +\Big(\frac{1}{2}\sum_{i=1}^{d-1}\pp{^2g}{x_i^2}(\v 0)\big(\kappa_i\frac{L^4}{5} + \sum_{j\neq i}\kappa_j\frac{L^4}{9}\big)- \frac{1}{2}\frac{L^2}3\sum^{d-1}_{i=1} \pp{^2g}{x_i^2}(\v 0) \frac{L^2}3\sum_{i=1}^{d-1}\kappa_i \Big) +\cO(L^6)\bigg)\\
	& = \pp{g}{y}(\v 0) + \frac{1}{2D}\bigg((\frac{L^4}{5}-\frac{L^4}{9})\sum_{i=1}^{d-1}\pp{^2g}{x_i^2}(\v 0)\kappa_i + \big(\sum_{i=1}^{d-1}\pp{^2g}{x_i^2}(\v 0)\sum_{j\neq i}\kappa_j - \sum_{i=1}^{d-1}\pp{^2g}{x_i^2}(\v 0)\sum_{j\neq i}\kappa_j\big)\frac{L^4}{9}+\cO(L^6)\bigg) \\
	& = \pp{g}{y}(\v 0) + \frac{1}{2D}\bigg(\frac{4L^4}{45}\sum_{i=1}^{d-1}\pp{^2g}{x_i^2}(\v 0)\kappa_i +\cO(L^6) \bigg)\\
	& = \pp{g}{y}(\v 0) + \frac{1}{2\sum_{i=1}^{d-1}\kappa_i^2\frac{4L^4}{45}}\bigg(\frac{4L^4}{45}\sum_{i=1}^{d-1}\pp{^2g}{x_i^2}(\v 0)\kappa_i +\cO(L^6) \bigg) \\
	& = \pp{g}{y}(\v 0) + \frac{1}{2}\frac{\displaystyle\sum_{i=1}^{d-1}\kappa_i \pp{^2g}{x_i^2}(\v 0)}{\displaystyle\sum_{i=1}^{d-1}\kappa_i^2} + \cO(L^2)
\end{align*}
\begin{align*}
	b^* &=\frac{1}{\expt{y^2} - \expt{y}^2}\bigg(g(\v 0)\Big(\expt{y^2} - \expt{y}^2\Big) +  \Big(\expt{y^2}\pp{g}{y}(\v 0)\expt{y} -\expt{y}\pp{g}{y}(\v 0)\expt{y^2}\Big) \\
	& \quad +\Big(\frac{1}{2}\expt{y^2}\sum_{i=1}^{d-1}\pp{^2g}{x_i^2}(\v 0)\expt{x_i^2} - \frac{1}{2}\expt{y}\sum^{d-1}_{i=1} \pp{^2g}{x_i^2}(\v 0) \expt{x_i^2y} \Big) +\cO(L^8)\bigg)\\
	&=g(\v 0) + \frac{1}{D}\bigg(\Big(\frac{1}{2}\sum_{i=1}^{d-1}\pp{^2g}{x_i^2}(\v 0)\frac{L^2}{3}\Big)\big(\sum_{i=1}^{d-1}\kappa_i^2\frac{L^4}{5} + 2\sum_{i=2}^{d-1}\sum_{j=1}^{i}\kappa_i\kappa_j\frac{L^4}{9}\big) \\
	 & \quad - \frac{1}{2}\frac{L^2}{3}\sum_{k=1}^{d-1}\kappa_k\sum^{d-1}_{i=1} \pp{^2g}{x_i^2}(\v 0)\big(\kappa_i\frac{L^4}{5} + \sum_{j\neq i}\kappa_j\frac{L^4}{9}\big) +\cO(L^8)\bigg)\\
	 & = g(\v 0) + \frac{1}{D}\bigg(\frac{1}{2}\sum_{i=1}^{d-1}\big(\sum_{j=1}^{d-1}\kappa_j^2\frac{L^6}{15} + 2\sum_{j=2}^{d-1}\sum_{k=1}^{j}\kappa_j\kappa_k\frac{L^6}{27}\big)\pp{^2g}{x_i^2}(\v 0)\\
	 & \quad - \frac{1}{2}\sum^{d-1}_{i=1} \Big(\big(\kappa_i^2\frac{L^6}{15} + \kappa_i\sum_{k\neq i}\kappa_k\frac{L^6}{15}
	 \big) + \big(\kappa_i \sum_{j\neq i}\kappa_j\frac{L^6}{27} +  (\sum_{j\neq i}\kappa_j)^2\frac{L^6}{27}\big)\Big)\pp{^2g}{x_i^2}(\v 0) +\cO(L^8)\bigg)\\
	 & = g(\v 0) + \frac{1}{D}\bigg(\frac{1}{2}\sum_{i=1}^{d-1}\big(\sum_{j=1}^{d-1}\kappa_j^2\frac{L^6}{15} + 2\sum_{j=2}^{d-1}\sum_{k=1}^{j}\kappa_j\kappa_k\frac{L^6}{27}\big)\pp{^2g}{x_i^2}(\v 0)\\
	 & \quad - \frac{1}{2}\sum^{d-1}_{i=1} \Big(\big(\kappa_i^2\frac{L^6}{15} + \kappa_i\sum_{k\neq i}\kappa_k\frac{L^6}{15}
	 \big) + \big(\kappa_i \sum_{j\neq i}\kappa_j +  \sum_{j\neq i}\kappa_j^2 + 2\sum_{\substack{j=2\\j\neq i}}^{d-1}\sum_{\substack{k=1\\k\neq i}}^{j} \kappa_j\kappa_k\big)\frac{L^6}{27}\Big)\pp{^2g}{x_i^2}(\v 0) +\cO(L^8)\bigg)\\
	 & = g(\v 0) + \frac{1}{2D}\sum_{i=1}^{d-1}\bigg(\sum_{j\neq i}\kappa_j^2(\frac{L^6}{15}-\frac{L^6}{27}) - (\frac{L^6}{15}+\frac{L^6}{27})\kappa_i\sum_{j\neq i}\kappa_j + 2\kappa_i\sum_{j\neq i}\kappa_j\frac{L^6}{27}\bigg)\pp{^2g}{x_i^2}(\v 0)+\cO(L^4) \\
	 & = g(\v 0) + \frac{1}{2D}\sum_{i=1}^{d-1}\bigg(\sum_{j\neq i}\kappa_j^2(\frac{L^6}{15}-\frac{L^6}{27}) - (\frac{L^6}{15}-\frac{L^6}{27})\kappa_i\sum_{j\neq i}\kappa_j \bigg)\pp{^2g}{x_i^2}(\v 0)+\cO(L^4)\\
	 & =  g(\v 0) + \frac{1}{2\displaystyle\sum_{i=1}^{d-1}\kappa_i^2\frac{4L^4}{45}}\sum_{i=1}^{d-1}\bigg(\sum_{j\neq i}\kappa_j^2(\frac{4L^6}{135}) - (\frac{4L^6}{135})\kappa_i\sum_{j\neq i}\kappa_j \bigg)\pp{^2g}{x_i^2}(\v 0)+\cO(L^4)\\
	 & = g(\v 0) + \frac{1}{2}\sum_{i=1}^{d-1} \pp{^2g}{x_i^2}(\v 0)\frac{L^2}{3}\Big(\frac{\displaystyle\sum_{j\neq i}\kappa_j^2 - \kappa_i\sum_{j\neq i}\kappa_j}{\displaystyle\sum_{k=1}^{d-1}\kappa_k^2}\Big) + \cO(L^4)
\end{align*}
\section{Local linear regression for curves in $\mathbb{R}^d$} \label{sec:curve}

In this section, we consider submanifold with codimension $d-1$ in some Euclidean space, that is, a smoothly embedded curve. A simple case would be a non-planar curve in $3$-space.
\subsection{Results on $3$-dimensional curves}
Similar as in \Cref{sec:hypersurface}, we fix a based point $\v x_0$ and apply unitary transformation and translation to obtain the standard local coordinate frame, where we have the tangent space $\cT_{\v x_0}\cM$ as our domain. However, the caveat here is that for a curve, the tangent space is always $1$-dimensional (line induced by the tangent vector), resulting in a normal space of $2$-dimensional. Since the principal curvatures are associated to a certain normal direction, when the normal space is only $1$-dimensional, there is no ambiguity, but when the codimension is larger than $1$, one needs to deal with care as we discuss in \Cref{sec:codim-k}.

Fortunately, for curves, a useful concept called Frenet–Serret frame from differential geometry comes in handy: if the curve $r$ is parameterized by the arc length $s$, then we have:
\begin{align*}
    \mathcal T(s) &= \frac{\v r'(s)}{|\v r'(s)|} &
    \mathcal N(s) &= \frac{\mathcal T'(s)}{|\mathcal T'(s)|}\\
    \mathcal B(s) &= \mathcal T(s) \times \mathcal N(s) &
    \kappa(s) &= |\mathcal T'(s)| = | \v r''(s)|
\end{align*}
where $\cT,\, \cN,\, \cB$ denotes the tangent direction, normal direction, and binormal direction respectively, and it is easy to see they form a orthonormal basis for the embedding space. $\kappa(s)$ denotes the curvature. Then the Frenet-Serret formulas give explicit equations to describe the relationship of the three basis vectors:
\begin{equation}\label{eqn:Frenet_3d}
	\begin{dcases}
		\cT'(s) = \kappa(s)\cN(s)\\
		\cN'(s) = -\kappa(s)\cT(s) + \tau(s)\cB(s)\\
		\cB'(s) = -\tau(s)\cN(s)
	\end{dcases}
\end{equation} where $\tau(s)$ is the torsion (at position $\v r(s)$) given by $\abs{\tau(s)} = \norm{\cN'(s)-\big(\cN'(s)\cdot\cT(s)\big)\cT(s)}$, which in intuition describes the tendency of the curve to deviate from being flat in a plane, just as the curvature describes the tendency to stay away from being straight as a line. Then we arrive at the following lemma.

\begin{lemma}\label{thm:3d_curve_representation}(Local representation of curves in $\bbR^3$)\\
For a smoothly embedded curve $\v r(s)$ in $3$-dimensional Euclidean space, locally when viewing from the coordinates defined through $\cT, \cN$ and $\cB$, it can be approximated by monomials in each of the basis direction in the following way:
\[  \v r(s) = \big(x, \frac{\kappa}2 x^2 + \cO(x^6), \frac{\kappa\tau}6 x^3 + \cO(x^9)\big) \implies \v r(s)\approx (x,  \frac{\kappa}2 x^2 , \frac{\kappa\tau}6 x^3)\]
\end{lemma}

\begin{proof}
By taking the local referring point as $\v r(0) = \v 0$, then a local Taylor expansion of the curve around $s=0$ yields:
\begin{equation} \label{eqn:curve_taylor_3d_incomplete}
	\v r(s) = \v 0 + \v r'(0) (s-0) + \frac{1}{2} \v r''(0) (s-0)^2 + \frac{1}{6} \v r'''(0)(s-0)^3 + \cO(s^4)
\end{equation}
Using the Frenet-Serret frame  \Cref{eqn:Frenet_3d}, we know
\begin{align*}
	\v r'(s) &= \cT(s)\\
	\v r''(s) &= \cT'(s) = \kappa(s)\cN(s)\\
	\v r'''(s) &= \big( \kappa(s)\cN(s)\big)' = \kappa'(s)\cN(s) + \kappa(s)N'(s)\\
	&= \kappa'(s)\cN(s) + \kappa(s) \big(-\kappa(s)\cT(s) + \tau(s)\cB(s)\big) \\
	&= \kappa'(s)\cN(s) -\kappa^2(s)\cT(s) + \kappa(s)\tau(s)\cB(s)
\end{align*}
Plug in the above into \Cref{eqn:curve_taylor_3d_incomplete} to obtain:
\begin{equation}\label{eqn:curve_taylor_3d}
	\v r(s) = \cT s + \frac{1}{2} \kappa\cN s^2 + \frac{1}{6} (\kappa'(0)\cN-\kappa^2\cT+\kappa\tau\cB)s^3+ \cO(s^4)
\end{equation}
 which is a complete characterization of the curve $\v r(s)$ around $s=0$ up to the forth order. Furthermore, note that $\cT$-$\cN$-$\cB$ forms an orthonormal basis with the origin being at $\v 0$, we can therefore express \Cref{eqn:curve_taylor_3d} up to the forth order term under the new coordinate basis by collecting terms as
 \[ \v r(s) \approx \Big(s - \frac{\kappa^2}{6} s^3,\; \frac{\kappa}{2}  s^2 +\frac{\kappa'(0)}{6} s^3,\; \frac{\kappa\tau}{6}  s^3\Big) \implies \v r(s) \approx \Big(s,\; \frac{\kappa}{2}  s^2,\; \frac{\kappa\tau}{6}  s^3\Big)\]
Finally, for another parametrization of $\v r(t)$, choose a specific $t$ such that the change along the tangent direction at $t=0$ has the same scale as the original $x$-axis. This can be easily done by setting the parametrization to have constant speed along $\cT$, and rescale the constant to match with $x$. Thus, locally we obtain $\v r(t)\cdot\cT=:x(t)\sim x$, having the same scale as $x$. 
On the other hand, for the arc length parametrization one has \[ s(t) = \int_0^t \norm{\v r'(z)}\dd{z}\] which is another function of $t$. Therefore, both $x(t)$ and $s(t)$ admit the Taylor series expansion around $0$ as follow: \begin{align*}
	s(t) &= s(0) + s'(0)t+ \frac{s''(0)}2t^2 + \cO(t^3)\\
	x(t) &= x(0) + x'(0)t+ \frac{x''(0)}2t^2 + \cO(t^3)
\end{align*} We show their Taylor series expansions match with each other in the lower order. Apparently $s(0) = x(0) = 0$, then for first order: $x'(t) = \v r'(t)\cdot\cT$, $s'(t) = \norm{\v r'(t)}$ by fundamental theorem of calculus. But $\cT = \displaystyle\frac{\v r'(0)}{\norm{\v r'(0)}}$, we have $x'(0) = \norm{\v r'(0)} = s'(0)$. As for the second order derivative, $x''(t) = \v r''(t)\cdot \cT\implies x''(0) = \v r''(0)\cdot \cT $.
\begin{align*}
	s''(t) &= (\norm{\v r'(t)})'
	=\frac{\inpdmid{\v r'(t)}{\v r''(t)}+\inpdmid{\v r''(t)}{\v r'(t)}}{2\sqrt{\inpdmid{\v r'(t)}{\v r'(t)}}}
	=\frac{\inpd{\v r'(t)}{\v r''(t)}}{\norm{\v r'(t)}} = \cT(t)\cdot \v r''(t)
\end{align*} which implies $s''(0) = \cT\cdot \v r''(0) = x''(0)$. A further calculation will show that $s'''(0)\neq x'''(0)$, therefore $s(t)-x(t) = \cO(t^3)$. Since $x(0) = 0$, from the Taylor expansion we have $x(t)=\cO(t)$, we can then write $s(t) = x(t) + \cO(x(t)^3)$. Plug this in to $\v r(s) \approx \Big(s,\; \frac{\kappa}{2}  s^2,\; \frac{\kappa\tau}{6}  s^3\Big)$  we have the desired results.
\end{proof}

The above formulation indicates that if we choose the tangent space as the space for the independent variable $x$, as we did in \Cref{sec:hypersurface}, and investigate the local graph representation of the curve with respect to this base coordinate, we have one graph along the $\cN$ direction, whose approximation is denoted by $y(x)= \frac{\kappa}{2}  x^2$, and one along the $\cB$ direction, with approximation denoted by  $z(x) = \frac{\kappa\tau}{6}  x^3$. Therefore, we can perform local linear regression under this coordinate frame, demonstrated in the following theorem:

\begin{theorem}\label{thm:3d_curve_sol} (Solution formulas for local linear regression on curves in $\bbR^3$)\\
Assume for simplicity the given data points $(x, y, z)\in \cM$ are uniform in $x$, $e.g.\;x\in\Omega = [-L,\, L]$ where $\cM$ is a curve in $3$D, then the solution formulas for local linear regression on $\cM$ under the local coordinate frame, if the problem is well-posed, are given by:
	\[\begin{dcases}
		w_x^* = \pp{g}{x}(\v 0) + \cO(L^4)\\
		w_y^* = \pp{g}{y}(\v 0)+\frac{1}{2k_2}\pp{^2g}{x^2}(\v 0) + \cO(L^2)\\
		w_z^* = \pp{g}{z}(\v 0) + \frac{k_2}{k_3} \pp{^2g}{xy}(\v 0) +\frac{1}{6k_3}\pp{^3g}{x^3}(\v 0) +\cO(L^2)\\
		b^* =  g(\v 0) + \cO(L^4)
\end{dcases}\]
where $k_2, k_3$ are the corresponding nonlinear quantities along the $y, z$ respectively.
\end{theorem}
\begin{proof}
From \Cref{thm:3d_curve_representation}, we know any curve  can be locally described by the triplet: $(x,\,\frac{\kappa}2 x^2,\, \frac{\kappa\tau}6 x^3)$ up to some higher order error, where $x$ is the independent variable along the tangent direction at the base point $\v x_0=\v 0$, and $\kappa,\,\tau$ are the curvature and torsion respectively at the base point. For consistency we denote $\kappa$ by $k_2$ and $\tau$ by $k_3$, and assume they are non-zero otherwise the linear regression problem becomes ill-posed on the dimensions with zero nonlinear quantities.

Similar to \Cref{sec:toy_model} by a symmetry argument, $e.g., \,\expt{xy}=\expt{yz}=0,\, \expt{xz}\neq0$, the linear system resulted from the least square minimization is (with a reordering):
\begin{equation*}
	\begin{bmatrix}
		\expt{x^2} &  \expt{xz} &0&0\\
		\expt{xz}&  \expt{z^2} &0& 0\\
		0& 0 & \expt{y^2}&  \expt{y}\\
		0& 0 & \expt{y}& 1
	\end{bmatrix}\begin{bmatrix}
		w_x \\ w_y \\ w_z\\ b
	\end{bmatrix}= \begin{bmatrix}
		\expt{gx} \\ \expt{gz} \\ \expt{gy} \\ \expt{g}
	\end{bmatrix}\implies \begin{dcases}
	\begin{bmatrix}
		\expt{x^2} &  \expt{xz}\\
		\expt{xz}&  \expt{z^2}\end{bmatrix}\begin{bmatrix}
			w_x \\ w_z\end{bmatrix} = \begin{bmatrix}
				\expt{gx} \\ \expt{gz}\end{bmatrix} \\
				\\
				\begin{bmatrix}
				 \expt{y^2}&  \expt{y}\\
					\expt{y}& 1\end{bmatrix}\begin{bmatrix}
					 w_y\\ b \end{bmatrix} = \begin{bmatrix}
							 \expt{gy} \\ \expt{g}\end{bmatrix}
\end{dcases},
\end{equation*}
leading to two independent $2\times2$ linear systems which can be solved exactly.  
\end{proof}
Again, from the solution formula we make conclusion similar to that from \Cref{sec:toy_model}: when any of the direction of $y$ and $z$ is flat, meaning $k_2=0$ or $k_3=0$, the linear system is not invertible hence the problem is well-posed; the non-linearity prevents the ill-posedness, but they affect the first order solution from a non-trivial way. The next step is to generalize  to curves in $d$-dimensional Euclidean space.

\subsection{Generalizations to curves in $\bbR^d$}
To obtain a similar result to curves in an Euclidean space of arbitrary dimension, we first need a local representation of the curve in $\bbR^d$. To this end, we introduce the notion of generalized Frenet-Serret formula: starting from the canonical basis vectors, the tangent $\cT$ and the normal $\cN(s) = \cT'(s)/\abs{\cT'(s)}$, one obtains the remaining orthonormal basis following a Gram-Schmidt type procedure by subtracting projection on previous directions. For example, to get the next two basis vectors:
 \[\tau\cB(s) = \cN'(s) - (\cN'(s)\cdot\cT(s))\cT(s) \implies\cN'(s)=-\kappa(s)\cT(s)+\tau(s)\cB(s)\]
\[\sigma\cD(s) = \cB'(s) -  (\cB'(s)\cdot\cN(s))\cN(s) - (\cB'(s)\cdot\cT(s))\cT(s)\implies \cB'(s)= -\tau\cN(s)+\sigma\cD(s),\]
where $(\tau,\,\cB)$ and $(\sigma,\,\cD)$ are the nonlinearity-direction pair for the third and the forth normal direction respectively, and the last equation follows from differentiating $\cB\cdot\cT = 0$. Last but not least, if the curve is embedded in $\bbR^4$, by expressing $\cD'(s)$ in terms of sum of projections along all the available directions, and differentiating the dot product of $\cD$ with all the directions, we will have $\displaystyle\cD'(s) = -\tau(s)\cB(s)$. In high dimension, for clarity, we use $\{\cV_i\}_{i=1}^d$ to denote the orthonormal basis, $e.g.\;\cV_1:=\cT$, $\cV_2:=\cN$, etc, and $\{\alpha_i\}_{i=1}^d$ for the corresponding curvature quantities, $e.g.\;\alpha_1 =1$, $\alpha_2=\kappa$, etc. Then, all the corresponding Frenet equations derived from the Gram-Schmidt process can be summarized in the following matrix form (with a slight abuse of notation that $\cV_1' = \cT'(0)$, etc.)

\begin{equation}\label{eqn:Frenet_nd}
	\begin{bmatrix}
		0 & \alpha_2 & 0   &\dots & 0\\
	-\alpha_2& 0   & \alpha_3  &\dots & 0\\
		0 & -\alpha_3 & 0  &\ddots & \vdots\\
		\vdots & \ddots  & \ddots & \ddots & \alpha_d\\
		0 & \dots & 0 & -\alpha_d & 0
	\end{bmatrix}\begin{bmatrix}
		\cV_1 \\ \cV_2 \\ \cV_3 \\ \vdots \\ \cV_d
\end{bmatrix}= \begin{bmatrix}
	 \cV_1' \\ \cV_2' \\ \cV_3' \\ \vdots \\ \cV_d'
\end{bmatrix}
\end{equation}
With \Cref{eqn:Frenet_nd}, we have the following Lemma:
\begin{lemma}\label{thm:nd_curve_representation}(Local representation of curves in $\bbR^d$)\\
For a smoothly embedded curve $\v r(s)$ in $\bbR^d$, locally when viewing from the coordinates frame obtained by the generalized Frenet-Serret frame, it can be approximated by monomials in each of the basis direction in the following way:
\[ \v r(s) \approx (s,\, \frac{\alpha_2}{2}s^2,\,\dots,\,\frac{\prod_{i=1}^d\alpha_i}{d!}s^d)\approx (x,\, k_2x^2,\dots,\, k_dx^d)\]
\end{lemma}
\begin{proof}
Note $\cV_1 = \v r'(0)$, and for $n<d$ from \Cref{eqn:Frenet_nd}, we notice:
\[\v r''(0) = \cV_1' = \alpha_2\cV_2,\; \v r'''(0) = \alpha_2\cV_2' =  \alpha_2(-\alpha_{2}\cV_1 + \alpha_{3}\cV_{3}), \; \cV_n' = -\alpha_{n}\cV_{n-1} + \alpha_{n+1}\cV_{n+1}\]
By induction, one can easily see that the $n$-th basis direction only shows up in and after the $n$-th order derivative $r^{(n)}(0)$. Combined with the Taylor series of $\v r$ around $s=0$, by setting $\v r(0) = 0$,
\[ \v r(s) = \sum_{n=1}^d \frac{\v r^{(n)}(0)}{n!}s^n + \cO(s^{d+1})\]
we know that when collecting all the terms associated to $\cV_n$, the leading order term is $s^n$. Therefore, the leading order approximation of $\v r$ in terms of the basis $\{\cV_i\}_{i=1}^d$ is:
\[ \v r(s) \approx (s,\, \frac{\alpha_2}{2}s^2,\,\dots,\,\frac{\prod_{i=1}^d\alpha_i}{d!}s^d)\]
 Using $x_n$ to denote the variable corresponding to the $n$-th direction and let $k_n := \frac{1}{n!}\prod_{i=1}^{n} \alpha_i$ to denote the corresponding nonlinear quantity, we get $x_n(s) = k_n s^n + \cO(s^{n+1})$. Then by the approximation introduced in similar to \Cref{thm:3d_curve_sol} , we finally arrive at the local approximation of the curve along the $n$-th direction for any $n\le d$:
\[ x_n(x) = k_n x^n + \cO(x^{3(n+1)}) \approx k_n x^n \]
\end{proof}
With \Cref{thm:nd_curve_representation}, one can follow the same procedure as in \Cref{thm:3d_curve_sol} to derive the solution formula for local linear regression on curves in $\bbR^d$. However, the resulting linear system is dense and high dimensional where the direct solving is no longer tractable. Therefore, we consider an approach based on method of matched asymptotic \cite{lagerstrom2013matched} to obtain the solution formulas.

\begin{theorem}\label{thm:nd_curve_sol}(Solution formulas for local linear regression on curves in $\bbR^d$)
Assume for simplicity the given data points $\v x\in \cM$ are uniform in $x_1$, $e.g.\;x_1\in\Omega = [-L,\, L]$ where $\cM$ is a curve in $\bbR^d$, then the solution formulas for local linear regression on $\cM$ under the local coordinate frame, if the problem is well-posed, are given by:
\[\begin{dcases}
   	w_n = \sum\limits_{\{j_i\}\in A_n}\frac{\prod\limits_{i=1} k_{j_i}}{|\{j_i\}|!k_n}\pp{^{|\{j_i\}|}g}{x_{j_1}x_{j_2}\dots}(\v 0) + \cO\Big(L^{2\big(\left\lceil\frac{d}{2}\right\rceil-\left\lfloor \frac{n}{2}\right\rfloor\big)}\Big)\\
	\,\\
  b =  g(\v 0)+ \cO\Big(L^{2\big(\left\lceil\frac{d}{2}\right\rceil\big)}\Big)
\end{dcases}.\]
where $k_j$ is the corresponding nonlinear quantities along the $j$-th basis direction, and $A_n$ is the set of all finite indexing sequences  $\left\{j_i\right\}_{i=1}$ such that:
\[ A_n = \left\{\, \{j_i\}_{i=1} \,\bigg|\, j_i \in N^*,\,\sum_i j_i = n \right\}\]
\end{theorem}
\begin{proof}
From \Cref{thm:nd_curve_representation}, variables associated to each basis direction can be expressed in terms of $x_1$, then by the symmetry argument: 
\[\expt{x_i x_j} \sim \expt{x_1^{i+j}} = \begin{dcases}
	\expt{x_1^{i+j}} = \frac{k_ik_j}{i+j+1} L^{i+j}; \; &\text{if } i+j \text{ even}\\
	0; \; &\text{if } i+j \text{ odd}
\end{dcases},\]
resulting in a decoupling of the linear system from the least square minimization of linear regression problem. For example, the odd system is given as 
\begin{equation}\label{eqn:lin_systm_curve_nd_odd}
	\begin{bmatrix}
		\expt{x_1^2} & \expt{x_1x_3}& \expt{x_1x_5}& \dots \\
		\expt{x_1x_3}& \expt{x_3^2}&\expt{x_3x_5}&\dots \\
		\expt{x_1x_5}& \expt{x_3x_5}&\expt{x_5^2}&\dots \\
		\vdots & \vdots & \vdots & \ddots \\
	\end{bmatrix}\begin{bmatrix}
		w_1 \\ w_3 \\ w_5 \\ \vdots
\end{bmatrix}= \begin{bmatrix}
	\expt{gx_1} \\ \expt{gx_3} \\ \expt{gx_5}\\ \vdots
\end{bmatrix},
\end{equation}
Solving the above system is not tractable in high dimension, however, there is a helpful one can make use of. For example, for the first row:
\[ \expt{x_1^2} = \frac{L^2}{3}; \quad \expt{x_1x_3} = \frac{k_3L^4}{5}; \quad \expt{x_1x_5} = \frac{k_5L^6}{7} \quad \dots \]
while for the R.H.S. $\; \expt{gx_1}$, assume $g$ has enough regularity, we Taylor expand $g$ around $\v 0$ abbreviating the evaluation at $\v 0$ notation for clarify $s.t.,\, \pp{g}{x_1} := \pp{g}{x_1}(\v 0)$:
\begin{align*}
  &\pp{g}{x_1}\expt{x_1^2} + \pp{g}{x_3}\expt{x_1x_3} + \pp{^2g}{x_1x_2}\expt{x_1^2x_2} +\pp{^2g}{x_2x_3}\expt{x_1x_2x_3} +\frac{1}6\pp{^3g}{x_1^3}\expt{x_1^4}+\frac{1}2\pp{^3g}{x_1^2x_3}\expt{x_1^3x_3} + \dots\\
	&= \pp{g}{x_1}\frac{L^2}{3} + \pp{g}{x_3}\frac{k_3L^4}{5} + \pp{^2g}{x_1x_2}\cO(L^4) +\pp{^2g}{x_2x_3}\cO(L^6) +\frac{1}6\pp{^3g}{x_1^3}L^4+\frac{1}2\pp{^3g}{x_1^2x_3}\cO(L^6) + \dots
\end{align*}
Since $L\ll 1$ is a free variable, by using the idea from the method of matched asymptotic expansions, we match different orders of $L$ in the equation to obtain the solutions for $w_i^*$. In this case, for the first three variable coefficeitns, we have 
\[\begin{dcases}
\frac{w_1^*}{3}L^2 = \pp{g}{x_1}\frac{L^2}{3} + \cO(L^4) \implies w_1^* = \pp{g}{x_1} + \cO(L^2)\\
 w_3^* = \pp{g}{x_3} + \frac{k_2}{k_3}\pp{^2g}{x_1x_2} + \frac{1}{6k_3}\pp{^3g}{x_1^3} + \cO(L^2) \\
 w_5^* = \pp{g}{x_5} + \frac{k_2k_3}{k_5}\pp{^2g}{x_2x_3} +  \frac{k_4}{k_5}\pp{^2g}{x_1x_4} +\frac{k_2^2}{2k_5}\pp{^3g}{x_1x_2^2} +\frac{k_3}{2k_5}\pp{^3g}{x_1^2x_3}+ \frac{k_2}{6k_5}\pp{^4g}{x_1^3x_2} + \frac{1}{120k_5}\pp{^5g}{x_1^5}   +\cO(L^2)
\end{dcases}\]
Last but not least, for $w_1^*$, the higher order $\cO(L^2)$ comes from the $\cO(L^4)$ terms from the R.H.S., but the existence of $w_3^*$ ensures the complete matching of the $\cO(L^4)$ terms, therefore, $w_1^*$ would have $\cO(L^4)$ as the higher order terms instead, but existence of $w_5^*$ would further turn it into $\cO(L^6)$. In the end, if the curve is embedded in $\bbR^d$ for $d$ being odd, the higher order term for $w_1^*$ would be $\cO(L^{d-1})$, and $\cO(L^d)$ if $d$ is even. Similar arguments can be applied to each of the coefficients. Finally, for the even order system for $b,\,w_2^*,\,w_4^*,\dots$, same conclusion can be made. And by identifying the patterns in the solution formulas, one could use the following compact formula to represent:
\[\begin{dcases}
   	w_n = \sum\limits_{\{j_i\}\in A_n}\frac{\prod\limits_{i=1} k_{j_i}}{|\{j_i\}|!k_n}\pp{^{|\{j_i\}|}g}{x_{j_1}x_{j_2}\dots}(\v 0) + \cO\Big(L^{2\big(\left\lceil\frac{d}{2}\right\rceil-\left\lfloor \frac{n}{2}\right\rfloor\big)}\Big)\\
	\,\\
  b =  g(\v 0)+ \cO\Big(L^{2\big(\left\lceil\frac{d}{2}\right\rceil\big)}\Big)
\end{dcases}.\]
where $k_j$ is the corresponding nonlinear quantities along the $j$-th basis direction, and $A_n$ is the set of all finite indexing sequences  $\left\{j_i\right\}_{i=1}$ such that:
\[ A_n = \left\{\, \{j_i\}_{i=1} \,\bigg|\, j_i \in N^*,\,\sum_i j_i = n \right\}.\]
For example, when $n=3$, $A_n = \left\{\,\{1,1,1\},\,\{1,2\}, \{2,1\}, \{3\}\,\right\}$, which are the corresponding $x$-indices for each of the term in the leading order solution for $w_3^*$.
\end{proof}

The above solution formulas indicate that when embedding the curve in an extremely high dimensional Euclidean space, the higher order term correspond to the power of $L$ would be infinitesimally small for most of the variable coefficients, while the leading order terms always remain, and we again observe the nonlinear quantities come into play in a totally non-trivial way.


\end{document}